\pdfoutput=1
\documentclass[11pt]{article}
\usepackage[final]{acl}

\usepackage{times}
\usepackage{latexsym}
\usepackage[T1]{fontenc}
\usepackage[utf8]{inputenc}
\usepackage{microtype}
\usepackage{inconsolata}
\usepackage{graphicx}
\usepackage{paralist}

\usepackage{xargs}
\usepackage{amssymb}
\usepackage{amsmath}
\usepackage{algorithm}
\usepackage{algpseudocode}
\usepackage{booktabs}
\usepackage{tabularx}
\usepackage{multirow}

\usepackage{pgfplots}
\usepackage{array}
\usepackage{multirow}
\usepackage{stfloats}
\usepackage{framed}
\usepackage{fancyvrb}
\usepackage{verbatimbox}
\usepackage{makecell}
\usepackage{lscape}
\usepackage{subcaption}
\usepackage{xspace}

\newcommand{\openness}{\textit{openness}\xspace}

\newcommand{\extraversion}{\textit{extraversion}\xspace}

\newcommand{\F}{$\textrm{F}_1$\xspace}
\newcommand{\RLP}{\textit{RL-Profiler}\xspace}
\newcommand{\RND}{\textit{RND}+\textit{CNet}\xspace}
\newcommand{\PMI}{\textit{PMI}+\textit{CNet}\xspace}
\newcommand{\ALL}{\textit{ALL}+\textit{CNet}\xspace}
\newcommand{\PT}{\textit{PT}+\textit{CNet}\xspace}
\newcommand{\baselineB}{\textit{Baseline-B}\xspace}
\newcommand{\baselineR}{\textit{Baseline-R}\xspace}

\newcommand{\PreserveBackslash}[1]{\let\temp=\\#1\let\\=\temp}
\newcolumntype{C}{>{\PreserveBackslash\centering}p{0.845cm}}
\newcolumntype{P}[1]{>{\raggedleft}p{#1}}
\newcolumntype{R}[1]{>{\raggedleft}p{#1}}
\newcolumntype{L}[1]{>{\raggedright}p{#1}}

\DeclareMathOperator{\sign}{sign}

\usetikzlibrary{shapes.geometric}
\pgfplotsset{compat=newest}
\pgfdeclareplotmark{starR}{
    \node[star,star point ratio=2.25,minimum size=9pt,
          inner sep=0pt,draw=black,solid,fill=yellow] {};
}

\newsavebox\CBox
\def\BF#1{\sbox\CBox{#1}\resizebox{\wd\CBox}{\ht\CBox}{\textbf{#1}}}

\newcommand{\widthGraphs}{1\textwidth}
\newcommand{\heightGraphs}{0.45\textwidth}

\renewcommand{\paragraph}[1]{\mbox{}\\[-0.8\baselineskip]\textbf{#1}}

\title{Prompt-based Personality Profiling:\\
  Reinforcement Learning for Relevance Filtering}

\author{
Jan Hofmann\textsuperscript{1},
Cornelia Sindermann\textsuperscript{2}, \and
Roman Klinger\textsuperscript{3}
\\
\textsuperscript{1}Institut f{\"u}r Maschinelle Sprachverarbeitung, University of Stuttgart, Germany \\
\textsuperscript{2}Computational Digital Psychology, Interchange Forum for Reflecting on\\ Intelligent Systems, University of Stuttgart, Germany\\
\textsuperscript{3}Fundamentals of Natural Language Processing, University of Bamberg, Germany \\
\texttt{\href{mailto:jan.hofmann@ims.uni-stuttgart.de}{jan.hofmann@ims.uni-stuttgart.de}};
  \texttt{\href{mailto:roman.klinger@uni-bamberg.de}{roman.klinger@uni-bamberg.de}}\\
  \texttt{\href{mailto:cornelia.sindermann@iris.uni-stuttgart.de}{cornelia.sindermann@iris.uni-stuttgart.de}}
}

\begin{document}
\maketitle
\begin{abstract}
  Author profiling is the task of inferring characteristics about
  individuals by analyzing content they share. Supervised machine
  learning still dominates automatic systems that perform this task,
  despite the popularity of prompting large language models to address
  natural language understanding tasks. One reason is that the
  classification instances consist of large amounts of posts,
  potentially a whole user profile, which may exceed the input length
  of Transformers. Even if a model can use a large context window, the
  entirety of posts makes the application of API-accessed black box
  systems costly and slow, next to issues which come with such
  ``needle-in-the-haystack'' tasks. To mitigate this limitation, we
  propose a new method for author profiling which aims at
  distinguishing relevant from irrelevant content first, followed by
  the actual user profiling only with relevant data. To circumvent the
  need for relevance-annotated data, we optimize this relevance filter
  via reinforcement learning with a reward function that utilizes the
  zero-shot capabilities of large language models.  We evaluate our
  method for Big Five personality trait prediction on two Twitter
  corpora. On publicly available real-world data with a skewed label
  distribution, our method shows similar efficacy to using all posts
  in a user profile, but with a substantially shorter context. An
  evaluation on a version of these data balanced with artificial posts
  shows that the filtering to relevant posts leads to a significantly
  improved accuracy of the predictions.
\end{abstract}
%
\section{Introduction}
Author profiling aims at inferring information about individuals by
analyzing content they share. A large and diverse set of
characteristics like age and gender \citep{Koppel2002,
  Argamon2003GenderGA, Schler2006EffectsOA}, native language
\citep{Koppel2005}, educational background \citep{coupland2007},
personality \citep{Pannebaker2003, Golbeck2011,
  kreuter-etal-2022-items}, or ideology \citep{Conover2011,
  PoliticEs2022} have been studied so far.  Author profiling is often
formulated supervised learning in which a full user profile with
possibly hundreds or thousands of individual textual instances
constitutes the input.

Despite the success of deep learning strategies in various natural
language processing tasks, such approaches often underperform when
applied to author profiling \citep{Lopez2023}. One factor contributing
to this may be that models like BERT \citep{Devlin2019} have
constraints on the length of the input they can process, preventing
them from processing all content linked to an author at once. Another
reason for this may be that not all content shared by an author is
equally useful when predicting certain characteristics. Some of the
content may even be considered noise, making it difficult for machine
learning models to grasp patterns needed when predicting specific
characteristics of an author -- we are faced with a
``needle-in-the-haystack''
challenge\footnote{\url{https://github.com/gkamradt/LLMTest_NeedleInAHaystack}}.

With this paper, we approach this challenge and propose to prefilter
posts to distinguish between helpful and misleading content before
inferring a characteristic. Thereby, accuracy of automated profiling
systems could be enhanced, and computational requirements could be
reduced. To induce such filter without data manually annotated for
relevancy, we study reinforcement learning with a reward function that
represents the expected performance gain of a prompt-based
system. Therefore, our approach only requires a prompt for a large
language model (LLM) and leads to a prefiltering classifier that can,
at test time, be applied with a limited number of queries to a large
language model. In contrast to retrieval augmented generation setups
\citep[RAG,][]{gao2024retrievalaugmentedgenerationlargelanguage}, our
setup has the advantage that it does not need to rely on the ad-hoc
abilities of a retrieval system.

Our contributions are therefore\footnote{The source code used in this
study is available at: \url{https://github.com/bluzukk/rl-profiler}}:
\begin{compactitem}
\item We propose a novel reinforcement learning-based relevance
  filtering method that we optimize with a reward inferred from the
  performance of a prompt-based zero-shot predictor.
\item We evaluate this method on personality prediction and show that
  a similar performance can be reached with limited, automatically
  filtered data, leading to a cheaper and environmentally more
  friendly social media analysis method.
\item We show the potential to improve the predictive performance with
  a partially artificial, balanced personality prediction corpus that
  we create via data augmentation. Here, the prediction is
  significantly more accurate with substantially smaller context.
\end{compactitem}

\section{Related Work}
\subsection{Zero-Shot Predictions with Large Language Models}
The terms prompt-based learning or in-context learning point at methods
in which we use an LLM's ability to generate text as a proxy for another
task. This approach has proven effective for a variety of tasks
\citep[i.a.]{Yin2019,gao-etal-2021-making,cui-etal-2021-template,ma-etal-2022-template,sainz-etal-2021-label,tu-etal-2022-prompt}.
For example, in a sentiment polarity classification, a
classification instance could be combined with a prompt that requests
a language model to output a word that corresponds either to a
positive or a negative class (``The food is very tasty.'' -- ``This
review is \underline{positive}/\underline{negative}.'').

State-of-the-art text classification
methods employ the Transformer architecture \citep{Vaswani2017}, which
are both deep and wide neural networks, optimized for parallel
processing of input data. However, they have a constrained input
length: BERT \citep{Devlin2019} can use 512 tokens, GPT-3.5 and Llama~2
\citep{Touvron2023} allow 4096 tokens, and GPT-4 \citep{GPT3} considers
8192 tokens\footnote{\url{https://agi-sphere.com/context-length/},
  access date 2024-07-22}. This situation makes the analysis of long texts
challenging and is the motivation for our work: automatically restricting
the data to be analyzed in a prompt to the most informative
segments.

One approach to solve this issue is to combine language-model based
text generation with information retrieval methods. In so-called RAG
(retrieval-augmented generation) approaches, the relevant passages for
a generation task are first retrieved in text-search manner, which are
then fed into the context of the language model
\citep{gao2024retrievalaugmentedgenerationlargelanguage}. In contrast
to our approach, such methods are optimized for ad-hoc retrieval, to
work with any given prompt.

\subsection{Personality in Psychology}
Stable patterns of characteristics and behaviors in individuals are
known as personality. Personality traits characterize differences
between individuals present over time and across situations.  Several
theories have been proposed attempting to categorize these differences
\citep[e.g.,][]{cattell1945, goldberg1981language,
  mccrae1992introduction}. Such theories include biologically oriented
ones \citep{cloninger_temperament_1994}, as well as lexical approaches
including the Five Factor Model \citep{digman_personality_1990} and
the \textsc{Hexaco} Model \citep{ashton_empirical_2007}.

\begin{table}
  \centering\small
  \begin{tabular}{lcc}
    \toprule
    I see Myself as Someone Who ... & Variable & Cor. \\
    \cmidrule(rl){1-1} \cmidrule(rl){2-2}\cmidrule(rl){3-3}
    ... does a thorough job & Consc. & $+$ \\
    ... can be somewhat careless & Consc. & $-$ \\
    ... is talkative & Extrav. & $+$ \\
    ... is reserved & Extrav. & $-$ \\
    ... worries a lot & Neurot. & $+$ \\
    ... is relaxed or handles stress well & Neurot. & $-$ \\
    \bottomrule
  \end{tabular}
  \caption{Example items from the BFI-44 questionnaire \citep{john1991big}. Negative scores indicate reversed-scored items.}
  \label{tab:examples_BFI44}
\end{table}

The Five Factor Model is one of the most extensively researched
and widely accepted models among personality psychologists, and proposes
that personality can be described based on five broad domains, the
so-called \textit{Big Five} of personality. Oftentimes, the Big Five
are named: \textit{openness to experience} (e.g., artistic, curious,
imaginative), \textit{conscientiousness} (e.g., efficient, organized,
reliable), \textit{extraversion} (e.g., active, outgoing, talkative),
\textit{agreeableness} (e.g., forgiving, generous, kind),
\textit{neuroticism} (e.g., anxious, unstable, worrying)
\citep{Costa1992}.  The Five Factor Model originates from the lexical
hypothesis stating that personality traits manifest in our language,
because we use it to describe human characteristics
\citep{brewer2019general, zis-Goldberg1990An, john1999big}.

A commonly used approach to assess the Big Five in individuals is the
application of self-report questionnaires, like the \textit{Big Five
  Inventory} (BFI) developed by \cite{john1991big}. This questionnaire
consists of 44 short phrases describing a person, and individuals are
asked to rate the extent to which they agree that each of these items
describes themselves on a five-point Likert scale from 1 (strongly
disagree) to 5 (strongly agree). Table~\ref{tab:examples_BFI44} shows
examples of these items. For example, if a person strongly agrees to
``being someone who is talkative'' and other related items of the same
scale, this can indicate a high level of \extraversion.

\begin{figure*}[t]
  \includegraphics[width=1\linewidth]{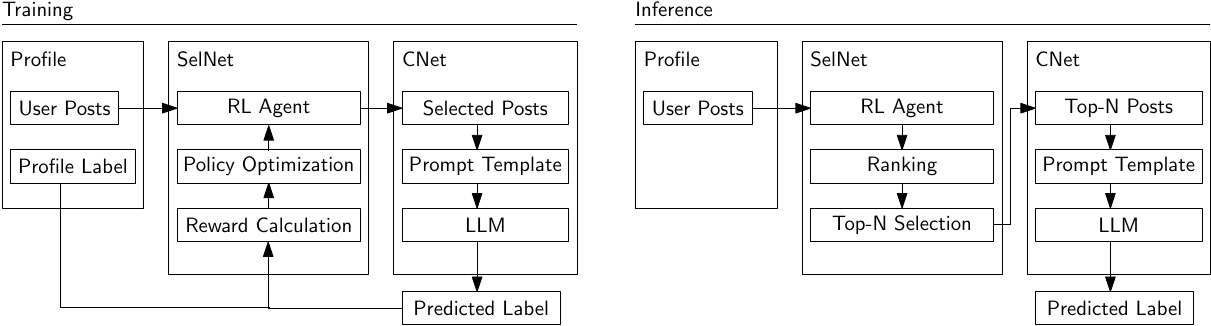}
  \caption{Overview on the workflow of the RL-Profiler (RL:
    Reinforcement Learning; SelNet: Selection Network; CNet:
    Classification Network; LLM: Large Language Model).}
  \label{fig:overview}
\end{figure*}

\subsection{Automatic Personality Prediction \newline from Text}
\label{sec:automatic}
One of the first attempts to personality prediction in social media
was proposed by \newcite{argamon2005lexical}, predicting
\textit{extraversion} and \textit{neuroticism} from essays on a binary
scale, i.e., predicting either a \textit{low} or \textit{high} level
of a trait.  Further, \newcite{schwartz2013personality} explored
written text on the social media platform Facebook, and found that
language use not only differs among people of different age and gender
but also among people rated differently along the Big Five traits. In
the 2015 PAN shared task \citep{rangel2015overview} the best results
predicting personality were obtained by \newcite{sulea2015automatic}
using ridge regression in combination with tf–idf weighted
character n-grams.

Since then, various deep learning approaches have been applied in
attempt to predict personality of users of social media platforms
\citep{khan2020personality}. These are, however, challenged by the
nature of the task: not all posts linked to individuals may be useful,
since content and tone of post from the same author may vary depending
on factors such as mood, current events, or specific interest at a
given time. Personality, however, characterizes differences between
persons present over time and across situations. Further, as not all
traits are strongly related to each other \citep{oz2015personality},
some posts might provide insights into one trait but not the
other. Consequently, there have been very limited efforts to predict
personality with the help of large language models
\citep{chinearios2022zerofewshotlearningauthor}. Accordingly, we argue
that systems would benefit from learning to differentiate between
relevant and misleading text instances by an author.

\section{RL-Profiler: Reinforcement Learning by LLM-based Performance Rewards}
We assume a profile consisting of a set of textual instances as input,
with annotations on the profile, but not instance level, during
training. We optimize the instance-relevance filter with information
from a profile-level prediction model. This filter decides which
textual instances are informative and should be used for the
profile-level decision.

Figure~\ref{fig:overview} illustrates this architecture. Our
\textbf{RL-Profiler} is devided into (1) the \textbf{Selection
  Network} (\textbf{SelNet}) and (2) the \textbf{Classification
  Network} (\textbf{CNet}).  SelNet corresponds to an agent in the RL
sense and selects textual instances from a profile. CNet then uses
these instances to predict a profile-level label. During training
(left side of Figure~\ref{fig:overview}), we compare this prediction
with the given profile-level ground truth to calculate a reward.

\subsection{Selection Network (SelNet)}
The core component of SelNet is the RL agent adopting a stochastic policy
$\pi(a \mid s, \theta)$ with the binary action space
$\mathcal{A} = \{\text{\textit{Select}}, \text{\textit{Reject}}\}$,
which we implement as a transformer-based classifier with a binary
classification head. Here, $\theta$ represents the trainable parameters,
$a \in \mathcal{A}$ denotes an action, and~$s$ is a single text instance
from a profile.

During training, an action is sampled from the probabilities given by
the agent's current policy. This ensures that the agent is exploring
different actions for the same input and the corresponding reward
during training. For inference, we adapt the behavior of SelNet: given
the set of instances from a profile, the policy of the trained agent is
first predicting probabilities for each instance. Then, all instances
are ranked by the predicted probability of
\textit{selecting} them and only the top-$N$ instances are fed to CNet
predicting a characteristic. This ensures that during inference,
the agent is no longer exploring different actions but only exploits
knowledge learned during training. Further, this forces SelNet to
always select a fixed number of instances $N$ from profiles,
eliminating the possibility of selecting no instance at all.

\subsubsection{Training the RL agent}

\begin{algorithm}[t!]
	\caption{RL-Profiler: Learning Algorithm}
	\label{alg:learning-algorithm}

	\begin{algorithmic}[1]
    \State \textbf{Input:} Policy $\pi_\theta$ with
    action space $\mathcal{A}=\{\textit{select}, \textit{reject}\}$,
    a training set $D$ with a set of profiles $\{P_1, ..., P_i\}$,
    each associated with a set of text instances $\mathcal{S}_P$
    and ground-truth $y_P$, and training epochs $E>0$

    \State Pre-train $\pi(a|s,\theta)$ using NPMI-Annotations

		\For{Epoch $i \leftarrow 1 \dots E$}
		\State Shuffle training set $D$
		\For{Profile $P$ in $D$}
		\State $C \leftarrow \{\}$ \Comment Set of selected instances.
		\For{Instance $s_t$ in $\mathcal{S}_P$}
    \State Sample action $a_t \in \{\textit{select}, \textit{reject}\}$ \phantom{xxxxxxxxx.} from $\pi(a_t | s_t, \theta)$
		\If{$a_t=\textit{select}$}
		\State $C \leftarrow C \cup s_t$
		\EndIf
		\EndFor
    \State $\hat{y}_P \leftarrow$ Prediction of CNet using $C$
    \State $R \leftarrow$ Reward using $y_P$, $\hat{y}_P$ and $C$
    \State \small$\theta \leftarrow \theta + \alpha \sum_{t=1}^{|\mathcal{S}_P|} (R-b) \ln \nabla_\theta \pi(a_t|s_t, \theta)$ \normalsize

		\EndFor
		\EndFor
	\end{algorithmic}
\end{algorithm}

Algorithm~\ref{alg:learning-algorithm} presents the method to train
the RL-agent. We use training data consisting of profiles with
associated ground-truth labels, and iterate multiple times over this
training dataset (Line~3).  In each epoch, profiles in the given
dataset are randomly arranged (Line~4). Given a single profile from
this training set, each instance from the profile is processed
individually (Line~7--12): the agent's current policy $\pi$ predicts a
probability for a single instance being relevant or irrelevant. In
other words, the agent predicts probabilities whether to select or
reject an instance.  During training, this action is sampled according
to the predicted probabilities (Line~8).  The selected text instances
are collected in a set $C$ (Line~10), and fed to CNet
predicting a profile-level label (Line~13). Using this prediction
and the ground-truth label, we then calculate a learning signal $R$
(cf.\ Equation~\ref{eq:reward}) to update the
policy of the agent (Line~14--15).

\paragraph{Reward.}
After collecting a subset of instances $C$ from a profile,
CNet uses this set to predict a label. We use this prediction
$\hat{y} \in \{0,1\}$, the ground-truth label $y \in \{0,1\}$
associated to the profile, and the number of selected instances
$|C|$ to calculate the reward~$R$:
\begin{equation}
	\label{eq:reward}
  R(y,\hat{y},C) = -2 + \sign(|C|) (3-2|y-\hat{y}|) - \lambda |C|
\end{equation}
with $\lambda$ being a hyperparameter that aims to decrease the reward
based on the number of selected instances. With this formulation of
the reward function, we summarize three cases: (1) if the predicted
label is equal to the ground-truth annotation we obtain
$+1 - \lambda |C|$, (2) if the predicted label is not equal to the
ground-truth annotation we obtain $-1 - \lambda |C|$, and (3) if the
set of selected posts is empty the reward is set to $-2$. Maximizing
this reward is the goal of the agent. Therefore, the agent needs to
learn to \textit{select} instances from profiles such that CNet
predicts the ground-truth label correctly, while \textit{rejecting} as
many instances as possible without rejecting all of them.

\paragraph{Policy Optimization.}
To optimize the behavior of the agent based on this reward, we adapt
the update rule of the \textsc{Reinforce} algorithm
\citep{Williams92}: given a profile $P$ associated with a set of text
instances $\mathcal{S}_P$, the parameters in $\theta$ are updated
based on the reward $R$ and the predicted probabilities of each of the
chosen actions following the current policy $\pi$:
\begin{equation}
  \label{eq:update-rule}
  \theta \leftarrow \theta + \alpha \sum_{t=1}^{|\mathcal{S}_P|} (R-b) \ln \nabla_\theta \pi(a_t|s_t, \theta),
\end{equation}

\noindent
where $b$ is a baseline. For simplicity, the calculation of $b$ is not shown
in Algorithm~\ref{alg:learning-algorithm}. In our approach we calculate
this baseline as the moving average reward given the last 10 update steps,
estimating the expected reward given the current policy.

\subsubsection{Pre-training using Mutual Information}
To improve stability of the training process of the RL agent \citep{mnih2015human},
we add a supervised pre-training step based on information theoretic measures
that associate words to labels.
We use normalized pointwise mutual information
\citep[NPMI,][]{bouma2009normalized,church-hanks-1990-word} to weigh
the association between each word $w$ present in text instances provided by a
profile and the corresponding ground-truth label $c$:
\begin{equation}
  \text{NPMI}(w;c) = \Bigr(\text{ln} \frac{p(w,c)}{p(w) p(c)}\Bigr) \Bigr / -\text{ln}~p(w, c)\,.
\end{equation}
We estimate these probabilities from the training set,
and use the NPMI weights to calculate a relevance-score
for individual instances. Here, for each instance $s \in \mathcal{S}_P$
associated to a profile $P$ we first calculate scores for each class $c$:
\begin{equation}
  \label{relevance-approx}
  \text{score}_c(s) = \sum_{w \in s} \text{NPMI}(w;c),
\end{equation}
and then a relevance-score considering all classes:
\begin{equation}
  \label{relevance-score}
  \text{r-score}(s, c_1, c_2) =
  \frac{\bigr|\text{score}_{c_1}(s)-\text{score}_{c_2}(s)\bigr|}{|\{w\mid
    w \in s \}|},
\end{equation}

\noindent
where $c_1$ and $c_2$ are the possible labels in a given author
profiling problem. Note that, for simplicity, we only consider binary
profile-level labels in this study (\textit{high} or \textit{low}),
and it is therefore sufficient to
define this score for two classes. After calculating a relevance-score
(r-score) for each text instance of all authors in the training set,
we annotate the top-$M$ instances of each author w.r.t.\ the highest
relevance-scores as \textit{relevant} while others are marked as
\textit{irrelevant}. These annotations are then used as a supervised
learning signal for pre-training the RL agent
(Line~2 in Algorithm~\ref{alg:learning-algorithm}).

\subsection{Classification Network (CNet)}
The combination of SelNet and CNet
forms a pipeline predicting a label given textual instances from a profile.
Given a set of selected text instances, CNet is responsible for predicting
this label. In this work, we propose to use a large language model in a
prompting setting for this purpose, since such a zero-shot setup does not
require any task specific training. Here, the classification task of predicting
a label from the selected text instances is verbalized, i.e., reformulated to
match the LLM's pre-training objective. CNet therefore creates a prompt using
the selected text instances by SelNet and a pre-defined prompt template.
We derive the classification result from the tokens the LLM generates in
response to such a prompt. The prompt setup is explained in the next section.

\section{Experiments}

\subsection{Experimental Setting and Training Details}
We implement RL-Profiler using the PyTorch \citep{paszke2019pytorch}
and HuggingFace's Transformer \citep{wolf2020huggingfaces} libraries.
For parameterizing the policy of the agent in SelNet, we use
\textit{bert-base-uncased}\footnote{\url{https://huggingface.co/google-bert/bert-base-uncased}},
and feed the \textit{[CLS]} token into a binary classification head
with a dropout \citep{srivastava2014dropout} of 20\%.  We pretrain the
agent using NPMI annotations marking the top-10 (top-$M$) instances as
relevant for 2 epochs, and fix the maximum epochs for reinforcement
learning to 200. During reinforcement learning, we fix $\lambda=.05$
for reward calculations, and adapt early stopping
by evaluating the current policy on validation data after each epoch
using different settings for top-$N$.  Here, we validate the current
policy by using the 5, 10, 20, 30, and 50 posts ($N \in \{5,10,20,30,50\}$)
of each profile the current policy predicts the highest probabilities
of selecting them. For each of these settings, we save the best model
checkpoint based on macro \F score. In both training phases
we use AdamW \citep{kingma2017adam, loshchilov2019decoupled}
with a learning rate of $10^{-6}$.

\begin{figure*}[t]
  \footnotesize
\begin{Verbatim}[frame=single]
<s>[INST] <<SYS>>
one word response
<</SYS>>

Recall the personality trait extraversion.
A person with a high level of extraversion may see themselves as someone who is talkative, or {...}
A person with a low level of extraversion may see themselves as someone who is reserved, or {...}

Consider the following tweets written by the same person:
{tweets}
Does this person show a low or high level of extraversion? Do not give an explanation. [/INST]
\end{Verbatim}
  \vspace{-0.5cm}
  \caption{Prompt template used in
    CNet for predicting a level of \extraversion.}
  \label{verb:prompt_template_llama2}
\end{figure*}
For the classification of the selected text instances (CNet) we use
\textit{Llama~2~13B-Chat}\footnote{\url{https://huggingface.co/TheBloke/Llama-2-13B-chat-GPTQ}}
\citep{touvron2023llama2} with GPTQ \citep{frantar2023gptq}, and fix
temperature to 0.8 and top-$p$ to 0.9 for all experiments. For all Big
Five traits we design individual prompts.
Figure~\ref{verb:prompt_template_llama2} show such a prompt for predicting
a level of extraversion.
Our prompts consist of a system prompt requesting
single word answers, context about a trait, the posts selected from
a profile, and an instruction. The context stems from items of the
BFI-44 \citep{john1991big} questionnaire used to score a particular
trait. These items are exemplarily added for a \textit{high} level
(``A person with a high level of extraversion may see themselves as
...''), while items that are scored in reversed are added as context
for a \textit{low} level (Table~\ref{tab:examples_BFI44} shows
examples of such items for other traits).

\subsection{Corpora}

We evaluate our approach on the English subset of the publicly
available PAN-AP-2015
data\footnote{\url{https://zenodo.org/records/3745945}}
\citep{rangel2015overview}. The personality trait annotations in this
corpus are derived from self-assessed BFI-10 online tests
\citep{rammstedt2007measuring}, a short version of the BFI-44. Here,
for each author, a score between $-$0.5 and 0.5 is provided for each
Big Five trait. We convert these scores to binary values at a
threshold of 0, and use 20\% of the training data for
validation for each trait, while ensuring a similar class distribution
in these sets. Note that this results in different dataset splits for
each trait. Table~\ref{table:PAN2015_statistics} summarizes the
statistics of the corpora we derive. On average over all traits and
splits, we find that each profile consists of 92.3 individual posts.

\begin{table}[t]
  \centering\small
  \begin{tabular}{lrrrrrrrrrrrrrrrrr}
    \toprule
    & \multicolumn{2}{c}{Training} & \multicolumn{2}{c}{Validation} & \multicolumn{2}{c}{Testing}\\
    \cmidrule(rl){2-3}\cmidrule(rl){4-5}\cmidrule(rl){6-7}
    Class  & \multicolumn{1}{c}{High}     & \multicolumn{1}{c}{Low}    & \multicolumn{1}{c}{High}     & \multicolumn{1}{c}{Low}      & \multicolumn{1}{c}{High} & \multicolumn{1}{c}{Low} \\
    \cmidrule(r){1-1}\cmidrule(rl){2-3}\cmidrule(rl){4-5}\cmidrule(rl){6-7}
    Open.  & 119                          & 1                          & 30                          & 1                            & 137                        & 1  \\
    Consc. & 93                           & 3                          & 24                          & 2                            & 113                        & 10 \\
    Extrav. & 96                           & 12                         & 24                          & 3                            & 114                        & 6  \\
    Agree. & 90                           & 15                         & 24                          & 4                            & 108                        & 11 \\
    Neurot. & 83                           & 30                         & 22                          & 8                            & 91                         & 39 \\
    \bottomrule
  \end{tabular}
  \caption{Corpus statistics of the splits derived from the
    PAN-AP-2015 \citep{rangel2015overview} corpus (in numbers of profiles).}
  \label{table:PAN2015_statistics}
\end{table}

\subsection{Baselines and Derived Systems}
We compare our method to two supervised-learning based approaches,
and four systems we directly derive from our method:

\paragraph{Baseline-R: Regression Classifier.}
For the first baseline, we adapt the best performing system
from the 2015 PAN shared task to fit the binary profiling problem.
In this system, \newcite{sulea2015automatic} use a ridge regression model
with character n-gram tf-idf features. We adapt this approach by
converting the ridge regressor into a ridge classifier.

\paragraph{Baseline-B: BERT Classifier.}
We adapt BERT (\textit{base-uncased}) to the binary classification problem
using a classification head.
Since the input to BERT is restricted to a maximum of 512 tokens,
all posts associated with an author can not be presented
to this model at once. Therefore, we propagate the profile-level
ground-truth to individual posts, and train BERT on post-level
for 2 epochs using a learning rate of $2\cdot10^{-5}$ with cross-entropy
loss weighted by class distribution. To obtain a profile-level prediction
we draw a majority vote from the predictions on individual posts.

\paragraph{ALL+CNet.}
We explore a variation of RL-Profiler that skips the selection
process of SelNet. In \ALL, all posts from
an author are given to CNet. Note that this is possible
in our experimental setting because the data we use only contains a
subset of posts from each user's profile.

\paragraph{RND+CNet.}
In this variation of RL-Profiler, we replace the reinforcement
learning agent in SelNet with a random selection of $N$ posts.

\paragraph{PMI+CNet.}
In this system, the selection process using the trained agent is replaced
by selecting~$N$ posts based on their \textit{relevance-score}
(Equation~\ref{relevance-score}). Here, instances are ranked
based on NPMI information and the top-$N$ instances are directly given to
CNet. With this system, we aim to provide insights on performance of such
a selection system when simply relying on information theoretic measures.

\paragraph{PT+CNet (Pre-train+CNet).}
Further, the agent trained using reinforcement learning can be replaced
by an agent that is only pre-trained on NPMI-Annotations, i.e., we
stop training the agent after Line~2 in Algorithm~\ref{alg:learning-algorithm}.

\subsection{Evaluation Procedure and Metrics}
We evaluate our experiments using macro-average and weighted-average
$\text{F}_1$ scores (average weighted by the number of instances per class).

Performance during evaluation of the individual systems in this study
can vary between runs. This is, for example, due to the
non-deterministic output generated by the LLM.  Therefore, we average
scores of 10 individual runs.

\subsection{Results}
\begin{table*}[t!]
  \centering\small
	\begin{tabular}{lrCCCCCCCCCCCCCCCCCCCCCCCCCCCCCCCC}
		\toprule
&& \multicolumn{2}{c}{Open.} & \multicolumn{2}{c}{Consc.} &\multicolumn{2}{c}{Extrav.} &\multicolumn{2}{c}{Agree.} &\multicolumn{2}{c}{Neur.} \\
\cmidrule(rl){3-4} \cmidrule(rl){5-6} \cmidrule(rl){7-8} \cmidrule(rl){9-10}\cmidrule(l){11-12}
System & top-$N$ & m-\F & w-\F & m-\F & w-\F & m-\F & w-\F & m-\F & w-\F & m-\F & w-\F\\
\cmidrule(r){1-1}\cmidrule(rl){2-2}\cmidrule(rl){3-4}\cmidrule(rl){5-6}\cmidrule(rl){7-8}\cmidrule(rl){9-10}\cmidrule(l){11-12}
\cmidrule(r){1-1}\cmidrule(rl){2-2}\cmidrule(rl){3-4}\cmidrule(rl){5-6}\cmidrule(rl){7-8}\cmidrule(rl){9-10}\cmidrule(l){11-12}
\baselineR & all & 49.8\scriptsize{$\pm$0.0}   & 98.9\scriptsize{$\pm$0.0}   & 47.9\scriptsize{$\pm$0.0}   & 88.0\scriptsize{$\pm$0.0}   & 82.5\scriptsize{$\pm$0.0}   & 96.7\scriptsize{$\pm$0.0}   & 59.8\scriptsize{$\pm$0.0}   & 88.2\scriptsize{$\pm$0.0}   & 66.8\scriptsize{$\pm$0.0}   & 73.6\scriptsize{$\pm$0.0}  \\
\baselineB & all & 56.5\scriptsize{$\pm$14.1}  & 99.0\scriptsize{$\pm$0.2}   & 58.8\scriptsize{$\pm$8.4}   & 88.8\scriptsize{$\pm$0.8}   & 76.9\scriptsize{$\pm$12.4}  & 93.6\scriptsize{$\pm$5.5}   & 66.2\scriptsize{$\pm$4.6}   & 88.2\scriptsize{$\pm$3.2}   & 67.8\scriptsize{$\pm$1.8}   & 73.6\scriptsize{$\pm$1.1}  \\
\ALL       & all & 49.8\scriptsize{$\pm$0.0}   & 98.9\scriptsize{$\pm$0.0}   & 47.0\scriptsize{$\pm$1.6}   & 70.5\scriptsize{$\pm$1.5}   & 48.4\scriptsize{$\pm$0.1}   & 92.1\scriptsize{$\pm$0.2}   & 52.5\scriptsize{$\pm$2.2}   & 75.8\scriptsize{$\pm$1.3}   & 42.7\scriptsize{$\pm$2.6}   & 57.0\scriptsize{$\pm$1.9}  \\
\cmidrule(r){1-1}\cmidrule(rl){2-2}\cmidrule(rl){3-4}\cmidrule(rl){5-6}\cmidrule(rl){7-8}\cmidrule(rl){9-10}\cmidrule(l){11-12}
\RLP & best & 47.7\scriptsize{$\pm$0.6}  & 94.6\scriptsize{$\pm$1.2}  & 44.6\scriptsize{$\pm$2.2}  & 63.9\scriptsize{$\pm$2.7}  & 57.0\scriptsize{$\pm$5.7}  & 92.3\scriptsize{$\pm$0.8}  & 43.1\scriptsize{$\pm$2.1}  & 70.8\scriptsize{$\pm$1.3}  & 39.3\scriptsize{$\pm$2.3}  & 47.0\scriptsize{$\pm$2.4} \\
\RND & best & 49.6\scriptsize{$\pm$0.1}  & 98.5\scriptsize{$\pm$0.3}  & 33.4\scriptsize{$\pm$1.9}  & 45.7\scriptsize{$\pm$3.0}  & 48.3\scriptsize{$\pm$0.2}  & 91.8\scriptsize{$\pm$0.3}  & 41.8\scriptsize{$\pm$2.0}  & 58.1\scriptsize{$\pm$1.9}  & 38.8\scriptsize{$\pm$1.7}  & 46.1\scriptsize{$\pm$0.8} \\
\PMI & best & 49.4\scriptsize{$\pm$0.2}  & 98.0\scriptsize{$\pm$0.3}  & 35.4\scriptsize{$\pm$1.9}  & 48.8\scriptsize{$\pm$2.9}  & 58.8\scriptsize{$\pm$3.8}  & 91.4\scriptsize{$\pm$1.1}  & 42.3\scriptsize{$\pm$1.6}  & 58.1\scriptsize{$\pm$1.9}  & 38.0\scriptsize{$\pm$1.6}  & 42.5\scriptsize{$\pm$1.7} \\
\PT  & best & 49.1\scriptsize{$\pm$0.3}  & 97.5\scriptsize{$\pm$0.6}  & 34.5\scriptsize{$\pm$1.9}  & 48.6\scriptsize{$\pm$1.9}  & 48.2\scriptsize{$\pm$0.2}  & 91.7\scriptsize{$\pm$0.4}  & 36.7\scriptsize{$\pm$2.2}  & 48.1\scriptsize{$\pm$2.8}  & 38.8\scriptsize{$\pm$1.7}  & 50.9\scriptsize{$\pm$1.6} \\
\toprule
	\end{tabular}
  \caption{Macro \F (m-\F) and weighted average \F (w-\F) scores for all models
    on testing data (average of 10 runs with standard deviation).
    For models with top-$N$ parameter (lower part in this table),
    the best setting based on macro \F score on validation data is chosen for
    each trait (validation results with all settings for top-$N$ shown in Table~\ref{tab:real_valid}).}
  \label{tab:results_real}
\end{table*}

In this section, we analyze the results of our experiments on the
PAN-AP-2015 corpus. To evaluate the effect of the number of selected
posts per profile we use validation data: for each trait we select the
setting for top-$N$ that produces the best results w.r.t.\ macro \F
score on validation data, individually for each method/baseline and
trait. We provide detailed results for all models and settings for
top-$N$ on validation data in the
Appendix~\ref{sec:appendix_validation_results}.

\paragraph{Does the prediction with partial data perform on par or
  worse in comparison to using all data?} Table~\ref{tab:results_real}
shows the results. Here, we are in particular interested whether our
approach is preferable compared to using all posts of profiles in a
zero-shot setting. We therefore compare the third and fourth row in
this Table and find that, for all traits except for \extraversion, our
approach (\RLP) performs only slightly worse compared to using all
posts (\ALL). On average over all traits, we find that \RLP performs
worse by 1.8pp macro \F (46.3\% vs. 48.1\%) and 5.2pp weighted \F
(78.9\% vs. 73.7\%). This is, although our method only uses 10 posts
from each profile on average over all traits while the \ALL system
uses 92.9.

\paragraph{Is RL-Profiler better than randomly selecting instances?}
One option to limit the amount of data is to choose a number of posts
at random. We therefore compare the fourth and fifth row in
Table~\ref{tab:results_real}, and observe that out method (\RLP) is
outperforming a random selection (\RND) for almost all traits (except
\openness, which has a majorly skewed class distribution).
On average over all traits, we find that our method
improves macro \F by 3.9pp (46.3\% vs. 42.4\%) and weighted \F by 5.7pp
(73.7\% vs. 68.0\%) compared to random selections. This is
although the \RND system is using $N$=50 posts, on four of the five traits,
while the proposed system only uses $N$=5 for these traits (since these
settings for $N$ produced the best results for these approaches during validation).

\paragraph{Is the RL necessary or would a purely statistical selection
suffice?}
This finding prompts the question of whether alternative selection methods that
bypass costly training could replace our trained RL agent. To explore this,
we compare our approach (Row 4) to its variants, \PMI (Row 6) and \PT (Row 7), and
observe that these alternatives generally underperform compared to the
trained agent,
Further, we compare \RLP to the two supervised learning-based systems \baselineR and \baselineB,
and find that, on average over all traits, performance decreases by 15.1pp
and 18.9pp macro \F, respectively, when using our zero-shot approach.

\paragraph{Computational Analysis.}
We perform our experiments on a single NVIDIA RTX A6000 (48GB) GPU with AMD
EPYC 7313 CPU and present the average prediction time per profile on testing data
for different zero-shot systems in Table~\ref{tab:runtime}.
For the \RLP and \RND systems, the reported time includes both the time
required to select a number of posts from each profile -- using the trained agent
or random selection, respectively -- and the time taken by CNet to generate a
prediction based on the selected posts. For the \ALL system this time only
reflects the duration required to retrieve a prediction from CNet.
We find that prediction time is substantially reduced by a reduced number of
selected posts. For example, when predicting extraversion, the average prediction
time for a profile is reduced by more than 76\% moving from 1.65s to 0.38s on the
comparison between our method and the system using all available posts in a zero-shot
setting.

\begin{table}[t]
	\centering\small
	\begin{tabular}{lrrrr}
		\toprule
    Variable & \RLP & \RND & \ALL \\
    \cmidrule(r){1-1}\cmidrule(rl){2-2}\cmidrule(rl){3-3}\cmidrule(rl){4-4}
    Open. & 0.54 (5)\phantom{0} & 1.11 (50) & 1.72 (94.3$^\star$)\\
    Consc. & 0.88 (5)\phantom{0}& 1.29 (50) & 2.11 (93.7$^\star$)\\
    Extrav. & 0.38 (5)\phantom{0}& 1.10 (50)& 1.65 (92.4$^\star$)\\
    Agree. & 0.61 (5)\phantom{0} & 1.12 (50)& 1.57 (91.9$^\star$)\\
    Neur. & 1.12 (30) & 1.03 (30) & 1.78 (92.1$^\star$)\\
		\toprule
	\end{tabular}
	\caption{Average prediction time in seconds per profile on testing data.
    For the \RLP and \RND system, the best setting for top-$N$ (in parentheses)
    based on validation performance is shown for each trait. For \ALL the number
    in parentheses denotes the average number of posts available per profile.}
	\label{tab:runtime}
\end{table}

\paragraph{Summary.}
We find that our approach is preferable to selecting data at
random when predicting personality, and only slightly worse
compared to using all available posts of profiles.
The advantage is that using only a small subset of posts increases
efficiency of the zero-shot setting drastically.

\subsection{Post-hoc Analysis with Artificial Data}
\begin{table*}[t]
  \centering\small
  \begin{tabular}{lrCCCCCCCCCCCCCCCCCCCCCCCCCCCCCCCC}
    \toprule
&& \multicolumn{2}{c}{Open.} & \multicolumn{2}{c}{Consc.} &\multicolumn{2}{c}{Extrav.} &\multicolumn{2}{c}{Agree.} &\multicolumn{2}{c}{Neur.} \\
\cmidrule(rl){3-4} \cmidrule(rl){5-6} \cmidrule(rl){7-8} \cmidrule(rl){9-10}\cmidrule(l){11-12}
System & top-$N$ & m-\F & w-\F & m-\F & w-\F & m-\F & w-\F & m-\F & w-\F & m-\F & w-\F\\
\cmidrule(r){1-1}\cmidrule(rl){2-2}\cmidrule(rl){3-4}\cmidrule(rl){5-6}\cmidrule(rl){7-8}\cmidrule(rl){9-10}\cmidrule(l){11-12}
\baselineR & all & 48.4\scriptsize{$\pm$0.0}   & 90.7\scriptsize{$\pm$0.0}   & 68.8\scriptsize{$\pm$0.0}   & 70.8\scriptsize{$\pm$0.0}   & 70.8\scriptsize{$\pm$0.0}   & 72.6\scriptsize{$\pm$0.0}   & 76.4\scriptsize{$\pm$0.0}   & 76.9\scriptsize{$\pm$0.0}   & 79.9\scriptsize{$\pm$0.0}   & 79.9\scriptsize{$\pm$0.0}  \\
\baselineB & all & 77.3\scriptsize{$\pm$21.2}  & 93.2\scriptsize{$\pm$8.3}   & 73.4\scriptsize{$\pm$6.0}   & 74.2\scriptsize{$\pm$5.8}   & 75.3\scriptsize{$\pm$16.7}  & 77.1\scriptsize{$\pm$18.2}  & 68.3\scriptsize{$\pm$9.5}   & 69.4\scriptsize{$\pm$9.0}   & 63.5\scriptsize{$\pm$6.7}   & 63.5\scriptsize{$\pm$6.7}  \\
\ALL       & all & 48.4\scriptsize{$\pm$0.0}   & 90.7\scriptsize{$\pm$0.0}   & 85.2\scriptsize{$\pm$4.0}   & 85.7\scriptsize{$\pm$3.9}   & 80.1\scriptsize{$\pm$6.9}   & 85.0\scriptsize{$\pm$4.9}   & 78.1\scriptsize{$\pm$2.0}   & 78.5\scriptsize{$\pm$2.0}   & 50.6\scriptsize{$\pm$3.8}   & 50.6\scriptsize{$\pm$3.8}  \\
\cmidrule(r){1-1}\cmidrule(rl){2-2}\cmidrule(rl){3-4}\cmidrule(rl){5-6}\cmidrule(rl){7-8}\cmidrule(rl){9-10}\cmidrule(l){11-12}
\RLP & best & 100.\scriptsize{$\pm$0.0}    & 100.\scriptsize{$\pm$0.0}  & 98.7\scriptsize{$\pm$2.1}  & 98.8\scriptsize{$\pm$1.9}  & 93.5\scriptsize{$\pm$4.8}  & 94.5\scriptsize{$\pm$4.2}  & 98.8\scriptsize{$\pm$1.9}  & 98.9\scriptsize{$\pm$1.8}  & 96.3\scriptsize{$\pm$1.1}  & 96.3\scriptsize{$\pm$1.1} \\
\RND & best & 48.4\scriptsize{$\pm$0.0}   & 90.7\scriptsize{$\pm$0.0}   & 69.0\scriptsize{$\pm$3.4}  & 68.7\scriptsize{$\pm$3.6}  & 45.6\scriptsize{$\pm$7.3}  & 60.9\scriptsize{$\pm$5.0}  & 71.3\scriptsize{$\pm$5.9}  & 71.0\scriptsize{$\pm$5.9}  & 43.1\scriptsize{$\pm$8.2}  & 43.1\scriptsize{$\pm$8.2} \\
\PMI & best & 45.2\scriptsize{$\pm$1.5}   & 84.7\scriptsize{$\pm$2.8}   & 81.7\scriptsize{$\pm$2.8}  & 82.2\scriptsize{$\pm$2.8}  & 69.6\scriptsize{$\pm$3.5}  & 77.6\scriptsize{$\pm$2.6}  & 67.4\scriptsize{$\pm$2.8}  & 66.6\scriptsize{$\pm$2.9}  & 72.9\scriptsize{$\pm$5.8}  & 72.9\scriptsize{$\pm$5.8} \\
\PT  & best & 46.6\scriptsize{$\pm$1.5}   & 87.4\scriptsize{$\pm$2.7}   & 67.6\scriptsize{$\pm$6.2}  & 66.9\scriptsize{$\pm$6.6}  & 52.1\scriptsize{$\pm$6.2}  & 65.5\scriptsize{$\pm$4.1}  & 81.1\scriptsize{$\pm$3.9}  & 81.0\scriptsize{$\pm$4.0}  & 62.9\scriptsize{$\pm$4.2}  & 62.9\scriptsize{$\pm$4.2} \\
\toprule
	\end{tabular}
  \caption{Macro \F (m-\F) and weighted average \F (w-\F) scores
    on artificially enriched testing data (average of 10 runs with standard deviation).
    For models with top-$N$ parameter (lower part),
    the best setting based on macro \F score on validation data is chosen for
    each trait (validation results with all settings for top-$N$ shown in Table~\ref{tab:enriched_valid}).}
  \label{tab:results_enriched}
\end{table*}
In the results we reported in the previous section we showed that
we obtain a similar zero-shot efficacy while improving
efficiency. There are presumably two major difficulties that lead to
the slight decrease in efficacy. Firstly, predictions on skewed
profile labels are notorously challenging. Secondly, it is not ensured
that every profile contains information that allows our agent to
learn. To evaluate the capabilities of our RL-Profiler approach, we
simplify the task by removing profiles of the majority classes and add
posts that ensure to express the personality trait of interest. This
is a reasonable analysis step, as the corpus we use is likely skewed
by the data acquisition procedures and does not represent the real
world distribution of personality traits in the population
\citep{kreuter-etal-2022-items}.

We therefore perform a post-hoc analysis on partially artificial data:
to ensure class distribution is fairly balanced, we select at most 15
profiles from training, validation and testing data for each class and
enrich all profiles with $\approx$5\% artificial posts we generate
using Llama 2. These artificially generated posts aim to clearly
indicate either a \textit{low} or \textit{high} level of a specific
trait, and we add such highly indicative posts to profiles based on
their ground-truth annotations. We present examples of artificially
generated posts, the process of generating such, and statistics about
this partially artificially corpus in the
Appendix~\ref{sec:appendix_data_generation}.

We repeat our experiments on this data and present the results in
Table~\ref{tab:results_enriched} (we present validation results in
the Appendix~\ref{sec:appendix_validation_results_artificial}).
In contrast to our previous experiments,
we find that our method majorly
outperforms the setting using all data (68.5\% vs.\ 97.5\% macro \F, +29pp on average over
all traits). In comparison to a random selection, we observe an even larger improvement
(53.5\% to 97.5\% macro \F, +44pp). Interestingly, on this data,
we find that our approach does not only outperform all zero-shot based
methods substantially, but also the supervised-learning based models:
compared to \baselineR and \baselineB, we observe an improvement of 28.6pp
and 25.9pp macro \F, respectively. These results indicate that our
method has large potential to improve needle-in-the-haystack
personality profiling tasks via prompting.

\section{Conclusion and Future Work}
We outlined a novel approach for automatic personality
prediction from social media data which enables prompt-based
predictions to focus on the most relevant parts of an input. Notably,
we do not require labels of relevance, but induce the filter only from
the prompt-performance on the profile level. While the results on real
data shows no performance improvement overall, it does decrease the
required context window of the language model. With an experiment on
artificial data, we can show a substantial performance
improvement. This shows that our method helps the language model to
focus on relevant content, instead of leaving this task to the
attention mechanisms in the transformer.

The present results provide several directions for future work: One
direction is to replace or adapt individual parts of the proposed
system. This includes the evaluation of other policy optimization
algorithms, exploring the usage of different large language models,
or experiment with different policy parameterization
techniques. Further, we suggest to study if the requirement for
labeled profiles could be relaxed by relying on confidence estimates
of the zero-shot classification.

Another interesting question would be if the relevancy assessement of
RL-Profiler is similar to what humans find relevant. This requires a
future annotation study of relevancy in personality
profiling. Finally, it also remains interesting to explore how our
approach performs when applied to predicting other concepts like
gender or age.

\section*{Acknowledgments}
This paper is supported by the project INPROMPT (Interactive Prompt
Optimization with the Human in the Loop for Natural Language Understanding
Model Development and Intervention, funded by the German Research
Foundation, KL 2869/13-1, project numer 521755488).
We acknowledge the support of the Ministerium f\"ur Wissenschaft,
Forschung und Kunst W\"urttemberg (MWK, Ministry of Science, Research
and the Arts Baden-W\"urttemberg under Az. 33-7533-9-19/54/5) in
``K\"unstliche Intelligenz \& Gesellschaft: Reflecting Intelligent
Systems for Diversity, Demography and Democracy (IRIS3D)'' and the
support by the ``Interchange Forum for Reflecting on Intelligent
Systems'' (IRIS) at the University of Stuttgart. Further, we
acknowledge the support by the Stuttgart Center for Simulation Science
(SimTech).

\section*{Ethical Considerations}
Personality profiling of social media users is an ethically
challenging task. We point out that all data we use stems from an
established data set, that has been, to the best of our knowledge,
collected following high ethical standards. We do not collect any data
ourselves. We condemn any applications of social media mining methods
applied to data of users who did not actively consent to using their
data for automatic processing. This is particularly the case for
subjective and imperfect prediction tasks in which the analysis may be
biased in a way that discriminates parts of a society, particularly
minority groups.

The methods we develop in this paper contribute to a more efficient
use of large language models, therefore contributing to a more
sustainable and resource-friendly use of computing
infrastructure. Nevertheless, automatic analysis methods need to be
applied with care, given the resources that they require.

\section*{Limitations}
While this study provides valuable insights, several limitations should
be acknowledged. First, we treated personality traits as binary variables.
However, personality is typically understood as a spectrum rather than a
binary value. This simplification potentially limits the applicability of
our findings to real-world scenarios where personality assessments are more
complex. Further, we did not evaluate our approach using very large-scale
language models. Performance of our approach with such models therefore
remains untested, and future research could explore how our method scales
with larger models to better understand its effectiveness.

Finally, due to resource-constraints, we did not perform exhaustive
hyperparameter optimization. This includes to allow different numbers
of instances for each profile to be considered. However, we did not
optimize them for one model more exhaustively than for
another. Therefore, we believe that this aspect would not change the
main results of our experiments.

\bibliography{custom}

\clearpage
\appendix

\onecolumn
\section{Appendix}
\label{sec:appendix}

\subsection{Validation Results}
\label{sec:appendix_validation_results}
\begin{table}[H]
  \centering\small
	\begin{tabular}{lrCCCCCCCCCCCCCCCCCCCCCCCCCCCCCCCC}
		\toprule
    && \multicolumn{2}{c}{Open.} & \multicolumn{2}{c}{Consc.} &\multicolumn{2}{c}{Extrav.} &\multicolumn{2}{c}{Agree.} &\multicolumn{2}{c}{Neur.} \\
		 \cmidrule(rl){3-4} \cmidrule(rl){5-6} \cmidrule(rl){7-8} \cmidrule(rl){9-10}\cmidrule(l){11-12}
    System & top-$N$ & m-\F & w-\F & m-\F & w-\F & m-\F & w-\F & m-\F & w-\F & m-\F & w-\F\\
\cmidrule(r){1-1}\cmidrule(rl){2-2}\cmidrule(rl){3-4}\cmidrule(rl){5-6}\cmidrule(rl){7-8}\cmidrule(rl){9-10}\cmidrule(l){11-12}
\RLP & 5 & \BF{55.4}\scriptsize{$\pm$10.9}  & \BF{92.0}\scriptsize{$\pm$1.7}  & \BF{49.5}\scriptsize{$\pm$5.4}  & \BF{70.6}\scriptsize{$\pm$3.9}  & \BF{49.6}\scriptsize{$\pm$8.0}  & \BF{82.5}\scriptsize{$\pm$2.3}  & \BF{70.1}\scriptsize{$\pm$3.2}  & \BF{84.4}\scriptsize{$\pm$2.5}  & 46.6\scriptsize{$\pm$6.0}  & 48.5\scriptsize{$\pm$5.9} \\
\RND & 5 & 45.9\scriptsize{$\pm$1.6}  & 88.8\scriptsize{$\pm$3.0}  & 32.4\scriptsize{$\pm$2.2}  & 43.5\scriptsize{$\pm$3.5}  & 45.8\scriptsize{$\pm$0.7}  & 81.4\scriptsize{$\pm$1.2}  & 44.4\scriptsize{$\pm$7.3}  & 53.6\scriptsize{$\pm$9.6}  & 37.6\scriptsize{$\pm$5.7}  & 36.0\scriptsize{$\pm$6.9} \\
\PMI & 5 & 44.4\scriptsize{$\pm$1.0}  & 86.0\scriptsize{$\pm$2.0}  & 30.5\scriptsize{$\pm$3.9}  & 40.6\scriptsize{$\pm$6.2}  & \BF{57.2}\scriptsize{$\pm$11.5}  & \BF{83.7}\scriptsize{$\pm$4.7}  & 47.3\scriptsize{$\pm$4.0}  & 57.8\scriptsize{$\pm$3.5}  & 29.6\scriptsize{$\pm$5.1}  & 27.8\scriptsize{$\pm$6.5} \\
\PT  & 5 & 45.4\scriptsize{$\pm$4.3}  & 85.6\scriptsize{$\pm$2.9}  & \BF{42.4}\scriptsize{$\pm$4.6}  & \BF{58.5}\scriptsize{$\pm$6.4}  & 45.5\scriptsize{$\pm$0.6}  & 80.8\scriptsize{$\pm$1.0}  & 45.5\scriptsize{$\pm$2.8}  & 52.8\scriptsize{$\pm$3.7}  & 29.6\scriptsize{$\pm$5.4}  & 27.6\scriptsize{$\pm$6.2} \\
\cmidrule(r){1-1}\cmidrule(rl){2-2}\cmidrule(rl){3-4}\cmidrule(rl){5-6}\cmidrule(rl){7-8}\cmidrule(rl){9-10}\cmidrule(l){11-12}
\RLP & 10 & 45.7\scriptsize{$\pm$1.1}  & 88.4\scriptsize{$\pm$2.2}  & 42.4\scriptsize{$\pm$6.1}  & 60.4\scriptsize{$\pm$6.3}  & 44.9\scriptsize{$\pm$1.2}  & 79.8\scriptsize{$\pm$2.1}  & 53.6\scriptsize{$\pm$6.7}  & 70.7\scriptsize{$\pm$5.3}  & 46.1\scriptsize{$\pm$4.4}  & 49.6\scriptsize{$\pm$4.0} \\
\RND & 10 & 45.5\scriptsize{$\pm$1.1}  & 88.1\scriptsize{$\pm$2.2}  & 26.5\scriptsize{$\pm$4.2}  & 34.0\scriptsize{$\pm$7.0}  & 44.6\scriptsize{$\pm$1.5}  & 79.3\scriptsize{$\pm$2.7}  & 39.6\scriptsize{$\pm$4.7}  & 47.3\scriptsize{$\pm$6.6}  & 41.3\scriptsize{$\pm$9.5}  & 40.4\scriptsize{$\pm$11.3} \\
\PMI & 10 & 44.8\scriptsize{$\pm$1.1}  & 86.8\scriptsize{$\pm$2.1}  & 23.5\scriptsize{$\pm$3.6}  & 29.1\scriptsize{$\pm$6.1}  & 50.5\scriptsize{$\pm$8.9}  & 80.2\scriptsize{$\pm$3.6}  & 45.2\scriptsize{$\pm$2.3}  & 52.5\scriptsize{$\pm$3.1}  & 42.8\scriptsize{$\pm$4.2}  & 43.9\scriptsize{$\pm$4.7} \\
\PT  & 10 & 43.5\scriptsize{$\pm$3.8}  & 82.3\scriptsize{$\pm$4.1}  & 32.9\scriptsize{$\pm$3.7}  & 44.3\scriptsize{$\pm$5.9}  & 44.4\scriptsize{$\pm$1.2}  & 79.0\scriptsize{$\pm$2.2}  & 37.8\scriptsize{$\pm$4.1}  & 42.4\scriptsize{$\pm$5.7}  & 33.5\scriptsize{$\pm$4.6}  & 34.1\scriptsize{$\pm$4.7} \\
\cmidrule(r){1-1}\cmidrule(rl){2-2}\cmidrule(rl){3-4}\cmidrule(rl){5-6}\cmidrule(rl){7-8}\cmidrule(rl){9-10}\cmidrule(l){11-12}
\RLP & 20 & 48.7\scriptsize{$\pm$0.6}  & 94.2\scriptsize{$\pm$1.2}  & 43.2\scriptsize{$\pm$2.2}  & 59.8\scriptsize{$\pm$3.0}  & 45.8\scriptsize{$\pm$0.9}  & 81.4\scriptsize{$\pm$1.5}  & 53.1\scriptsize{$\pm$4.8}  & 71.8\scriptsize{$\pm$2.8}  & 42.6\scriptsize{$\pm$4.8}  & 49.6\scriptsize{$\pm$4.4} \\
\RND & 20 & 47.5\scriptsize{$\pm$0.5}  & 92.0\scriptsize{$\pm$1.0}  & 29.0\scriptsize{$\pm$4.0}  & 38.1\scriptsize{$\pm$6.4}  & 45.4\scriptsize{$\pm$0.8}  & 80.8\scriptsize{$\pm$1.4}  & 41.8\scriptsize{$\pm$7.4}  & 50.9\scriptsize{$\pm$9.8}  & 42.7\scriptsize{$\pm$5.2}  & 47.9\scriptsize{$\pm$4.9} \\
\PMI & 20 & 46.5\scriptsize{$\pm$0.9}  & 90.1\scriptsize{$\pm$1.7}  & 25.9\scriptsize{$\pm$1.5}  & 33.1\scriptsize{$\pm$2.5}  & 43.0\scriptsize{$\pm$1.2}  & 76.5\scriptsize{$\pm$2.1}  & 39.7\scriptsize{$\pm$4.2}  & 46.7\scriptsize{$\pm$5.2}  & \BF{45.1}\scriptsize{$\pm$2.3}  & \BF{50.2}\scriptsize{$\pm$2.6} \\
\PT  & 20 & 46.3\scriptsize{$\pm$1.0}  & 89.5\scriptsize{$\pm$1.9}  & 30.2\scriptsize{$\pm$4.1}  & 40.0\scriptsize{$\pm$6.7}  & 44.8\scriptsize{$\pm$0.8}  & 79.6\scriptsize{$\pm$1.5}  & 45.0\scriptsize{$\pm$4.4}  & 54.3\scriptsize{$\pm$5.4}  & 41.0\scriptsize{$\pm$3.2}  & 47.1\scriptsize{$\pm$3.3} \\
\cmidrule(r){1-1}\cmidrule(rl){2-2}\cmidrule(rl){3-4}\cmidrule(rl){5-6}\cmidrule(rl){7-8}\cmidrule(rl){9-10}\cmidrule(l){11-12}
\RLP & 30 & 48.9\scriptsize{$\pm$0.4}  & 94.7\scriptsize{$\pm$0.8}  & 40.2\scriptsize{$\pm$4.4}  & 56.0\scriptsize{$\pm$5.2}  & 47.1\scriptsize{$\pm$0.0}  & 83.7\scriptsize{$\pm$0.0}  & 51.5\scriptsize{$\pm$2.9}  & 68.0\scriptsize{$\pm$2.8}  & \BF{48.8}\scriptsize{$\pm$5.1}  & \BF{56.6}\scriptsize{$\pm$4.2} \\
\RND & 30 & 47.8\scriptsize{$\pm$1.2}  & 92.5\scriptsize{$\pm$2.3}  & 30.0\scriptsize{$\pm$2.9}  & 39.7\scriptsize{$\pm$4.6}  & 46.1\scriptsize{$\pm$1.1}  & 81.9\scriptsize{$\pm$1.9}  & 43.5\scriptsize{$\pm$6.1}  & 53.8\scriptsize{$\pm$7.9}  & \BF{44.1}\scriptsize{$\pm$4.6}  & \BF{51.7}\scriptsize{$\pm$3.7} \\
\PMI & 30 & 46.2\scriptsize{$\pm$0.8}  & 89.4\scriptsize{$\pm$1.5}  & 30.0\scriptsize{$\pm$2.5}  & 39.7\scriptsize{$\pm$4.0}  & 44.8\scriptsize{$\pm$1.3}  & 79.6\scriptsize{$\pm$2.4}  & 47.5\scriptsize{$\pm$5.2}  & 58.0\scriptsize{$\pm$6.0}  & 38.2\scriptsize{$\pm$3.0}  & 47.4\scriptsize{$\pm$3.1} \\
\PT  & 30 & \BF{47.7}\scriptsize{$\pm$0.7}  & \BF{92.3}\scriptsize{$\pm$1.4}  & 33.5\scriptsize{$\pm$2.8}  & 45.3\scriptsize{$\pm$4.5}  & 45.1\scriptsize{$\pm$0.5}  & 80.2\scriptsize{$\pm$0.8}  & \BF{47.5}\scriptsize{$\pm$4.6}  & \BF{57.2}\scriptsize{$\pm$4.5}  & 40.9\scriptsize{$\pm$6.8}  & 48.6\scriptsize{$\pm$5.5} \\
\cmidrule(r){1-1}\cmidrule(rl){2-2}\cmidrule(rl){3-4}\cmidrule(rl){5-6}\cmidrule(rl){7-8}\cmidrule(rl){9-10}\cmidrule(l){11-12}
\RLP & 50 & 48.7\scriptsize{$\pm$0.6}  & 94.2\scriptsize{$\pm$1.2}  & 44.5\scriptsize{$\pm$2.9}  & 61.6\scriptsize{$\pm$3.8}  & 47.1\scriptsize{$\pm$0.0}  & 83.7\scriptsize{$\pm$0.0}  & 50.8\scriptsize{$\pm$4.3}  & 68.2\scriptsize{$\pm$2.9}  & 45.6\scriptsize{$\pm$5.9}  & 58.5\scriptsize{$\pm$4.2} \\
\RND & 50 & \BF{48.4}\scriptsize{$\pm$0.8}  & \BF{93.7}\scriptsize{$\pm$1.5}  & \BF{36.9}\scriptsize{$\pm$4.1}  & \BF{50.5}\scriptsize{$\pm$6.1}  & \BF{47.1}\scriptsize{$\pm$0.0}  & \BF{83.7}\scriptsize{$\pm$0.0}  & \BF{47.4}\scriptsize{$\pm$5.2}  & \BF{62.3}\scriptsize{$\pm$7.0}  & 42.0\scriptsize{$\pm$6.0}  & 56.9\scriptsize{$\pm$4.6} \\
\PMI & 50 & \BF{48.3}\scriptsize{$\pm$0.4}  & \BF{93.5}\scriptsize{$\pm$0.8}  & \BF{34.1}\scriptsize{$\pm$1.9}  & \BF{46.3}\scriptsize{$\pm$3.0}  & 47.1\scriptsize{$\pm$0.0}  & 83.7\scriptsize{$\pm$0.0}  & \BF{51.6}\scriptsize{$\pm$3.3}  & \BF{65.9}\scriptsize{$\pm$3.0}  & 39.0\scriptsize{$\pm$4.8}  & 54.6\scriptsize{$\pm$4.0} \\
\PT  & 50 & 47.7\scriptsize{$\pm$0.6}  & 92.4\scriptsize{$\pm$1.2}  & 37.5\scriptsize{$\pm$3.8}  & 51.3\scriptsize{$\pm$5.7}  & \BF{46.7}\scriptsize{$\pm$0.5}  & \BF{83.1}\scriptsize{$\pm$0.9}  & 44.8\scriptsize{$\pm$5.3}  & 59.0\scriptsize{$\pm$5.7}  & \BF{44.9}\scriptsize{$\pm$6.7}  & \BF{58.9}\scriptsize{$\pm$5.5} \\
\cmidrule(r){1-1}\cmidrule(rl){2-2}\cmidrule(rl){3-4}\cmidrule(rl){5-6}\cmidrule(rl){7-8}\cmidrule(rl){9-10}\cmidrule(l){11-12}
\ALL & all & 49.2\scriptsize{$\pm$0.0}  & 95.2\scriptsize{$\pm$0.0}  & 41.6\scriptsize{$\pm$2.8}  & 65.6\scriptsize{$\pm$2.2}  & 47.1\scriptsize{$\pm$0.0}  & 83.7\scriptsize{$\pm$0.0}  & 54.8\scriptsize{$\pm$3.9}  & 76.5\scriptsize{$\pm$2.7}  & 42.7\scriptsize{$\pm$5.1}  & 60.1\scriptsize{$\pm$3.5} \\
\toprule
	\end{tabular}
  \caption{\small{Macro \F (m-\F) and weighted average \F (w-\F) scores for
    selection-based models with different settings for the top-$N$ hyperparameter
    on validation data (averages of 10 runs with standard derivation).
    The best performing setting for top-$N$ (w.r.t. the highest m-\F)
    for each model and personality trait (highlighted in \BF{bold}) is selected
    for evaluation on testing data.}}
  \label{tab:real_valid}
\end{table}
\vspace{-0.25cm}
\begin{figure}[H]
\centering
\begin{subfigure}{0.5\textwidth}
\pgfplotstableread[row sep=\\,col sep=&]{
  interval & RL    & RND  & PMI   & PRE &R&B\\
5   & 55.4 & 45.9 & 44.4 & 45.4&&\\
10   & 45.7 & 45.5 & 44.8 & 43.5&&\\
20   & 48.7 & 47.5 & 46.5 & 46.3&&\\
30   & 48.9 & 47.8 & 46.2 & 47.7&&\\
50   & 48.7 & 48.4 & 48.3 & 47.7&&\\
All  & 49.2 & 49.2 & 49.2 & 49.2 && 49.2\\
}\mydatax

  \caption*{Openness}
  \vspace{-0.2cm}
\footnotesize
\begin{tikzpicture}
    \begin{axis}[
            width=\widthGraphs,
            height=\heightGraphs,
            legend style={at={(1.32,0)},anchor=south east},
            symbolic x coords={5,10,.,20,.,30,.,.,50,.,.,All},
            ymajorgrids=true,
            xtick=data,
            every axis plot/.append style={semithick},
            nodes near coords align={vertical},
            ymin=20,ymax=80,
            y label style={at={(-0.075,0.5)}},
            x label style={at={(0.5,-0.125)}},
            xtick pos=bottom
        ]
\addplot table[x=interval,y=RL]{\mydatax};
\addplot table[x=interval,y=RND]{\mydatax};
\addplot table[x=interval,y=PMI]{\mydatax};
\addplot table[x=interval,y=PRE]{\mydatax};
\addplot[only marks, mark=starR, black] table[x=interval,y=R]{\mydatax};
\addplot[only marks, mark=starR, black] table[x=interval,y=B]{\mydatax};
\end{axis}
\end{tikzpicture}
\end{subfigure}
\hspace{-0.35cm}
\begin{subfigure}[b]{0.5\textwidth}
\pgfplotstableread[row sep=\\,col sep=&]{
  interval & RL    & RND  & PMI   & PRE &R&B\\
5   & 49.5 & 32.4 & 30.5 & 42.4&&\\
10   & 42.4 & 26.5 & 23.5 & 32.9&&\\
20   & 43.2 & 29.0 & 25.9 & 30.2&&\\
30   & 40.2 & 30.0 & 30.0 & 33.5&&\\
50   & 44.5 & 36.9 & 34.1 & 37.5&&\\
All  & 41.6 & 41.6 & 41.6 & 41.6 && 41.6\\
    }\mydata

  \caption*{Conscientiousness}
\vspace{-0.2cm}
\footnotesize
\begin{tikzpicture}
    \begin{axis}[
            width=\widthGraphs,
            height=\heightGraphs,
            legend style={at={(1.32,0)},anchor=south east},
            symbolic x coords={5,10,.,20,.,30,.,.,50,.,.,All},
            ymajorgrids=true,
            xtick=data,
            every axis plot/.append style={semithick},
            nodes near coords align={vertical},
            ymin=20,ymax=80,
            y label style={at={(-0.075,0.5)}},
            x label style={at={(0.5,-0.15)}},
            xtick pos=bottom
        ]
\addplot table[x=interval,y=RL]{\mydata};
\addplot table[x=interval,y=RND]{\mydata};
\addplot table[x=interval,y=PMI]{\mydata};
\addplot table[x=interval,y=PRE]{\mydata};
\addplot[only marks, mark=starR, black] table[x=interval,y=R]{\mydata};
\addplot[only marks, mark=starR, black] table[x=interval,y=B]{\mydata};
\end{axis}
\end{tikzpicture}
\end{subfigure}

\begin{subfigure}{0.5\textwidth}
\pgfplotstableread[row sep=\\,col sep=&]{
  interval & RL    & RND  & PMI   & PRE &R&B\\
5   & 49.6 & 45.8 & 57.2 & 45.5&&\\
10   & 44.9 & 44.6 & 50.5 & 44.4&&\\
20   & 45.8 & 45.4 & 43.0 & 44.8&&\\
30   & 47.1 & 46.1 & 44.8 & 45.1&&\\
50   & 47.1 & 47.1 & 47.1 & 46.7&&\\
All  & 47.1 & 47.1 & 47.1 & 47.1 && 47.1\\
}\mydatax

  \caption*{Extraversion}
  \vspace{-0.2cm}
\footnotesize
\begin{tikzpicture}
    \begin{axis}[
            width=\widthGraphs,
            height=\heightGraphs,
            legend style={at={(1.32,0)},anchor=south east},
            symbolic x coords={5,10,.,20,.,30,.,.,50,.,.,All},
            ymajorgrids=true,
            xtick=data,
            every axis plot/.append style={semithick},
            nodes near coords align={vertical},
            ymin=20,ymax=80,
            y label style={at={(-0.075,0.5)}},
            x label style={at={(0.5,-0.125)}},
            xtick pos=bottom
        ]
\addplot table[x=interval,y=RL]{\mydatax};
\addplot table[x=interval,y=RND]{\mydatax};
\addplot table[x=interval,y=PMI]{\mydatax};
\addplot table[x=interval,y=PRE]{\mydatax};
\addplot[only marks, mark=starR, black] table[x=interval,y=R]{\mydatax};
\addplot[only marks, mark=starR, black] table[x=interval,y=B]{\mydatax};
\end{axis}
\end{tikzpicture}
\end{subfigure}
\hspace{-0.35cm}
\begin{subfigure}[b]{0.5\textwidth}
\pgfplotstableread[row sep=\\,col sep=&]{
  interval & RL    & RND  & PMI   & PRE &R&B\\
5   & 70.1 & 44.4 & 47.3 & 45.5&&\\
10   & 53.6 & 39.6 & 45.2 & 37.8&&\\
20   & 53.1 & 41.8 & 39.7 & 45.0&&\\
30   & 51.5 & 43.5 & 47.5 & 47.5&&\\
50   & 50.8 & 47.4 & 51.6 & 44.8&&\\
All  & 54.8 & 54.8 & 54.8 & 54.8 && 54.8\\
}\mydata

  \caption*{Agreeableness}
\vspace{-0.2cm}
\footnotesize
\begin{tikzpicture}
    \begin{axis}[
            width=\widthGraphs,
            height=\heightGraphs,
            legend style={at={(1.32,0)},anchor=south east},
            symbolic x coords={5,10,.,20,.,30,.,.,50,.,.,All},
            ymajorgrids=true,
            xtick=data,
            every axis plot/.append style={semithick},
            nodes near coords align={vertical},
            ymin=20,ymax=80,
            y label style={at={(-0.075,0.5)}},
            x label style={at={(0.5,-0.15)}},
            xtick pos=bottom
        ]
\addplot table[x=interval,y=RL]{\mydata};
\addplot table[x=interval,y=RND]{\mydata};
\addplot table[x=interval,y=PMI]{\mydata};
\addplot table[x=interval,y=PRE]{\mydata};
\addplot[only marks, mark=starR, black] table[x=interval,y=R]{\mydata};
\addplot[only marks, mark=starR, black] table[x=interval,y=B]{\mydata};
\end{axis}
\end{tikzpicture}
\end{subfigure}

\begin{subfigure}{0.5\textwidth}
\pgfplotstableread[row sep=\\,col sep=&]{
  interval & RL    & RND  & PMI   & PRE &R&B\\
5   & 46.6 & 37.6 & 29.6 & 29.6&&\\
10   & 46.1 & 41.3 & 42.8 & 33.5&&\\
20   & 42.6 & 42.7 & 45.1 & 41.0&&\\
30   & 48.8 & 44.1 & 38.2 & 40.9&&\\
50   & 45.6 & 42.0 & 39.0 & 44.9&&\\
All  & 42.7 & 42.7 & 42.7 & 42.7 && 42.7\\
}\mydatax

  \caption*{Neuroticism}
  \vspace{-0.2cm}
\footnotesize
\begin{tikzpicture}
    \begin{axis}[
            width=\widthGraphs,
            height=\heightGraphs,
            legend style={at={(1.61,0)},anchor=south east},
            symbolic x coords={5,10,.,20,.,30,.,.,50,.,.,All},
            ymajorgrids=true,
            xtick=data,
            every axis plot/.append style={semithick},
            nodes near coords align={vertical},
            ymin=20,ymax=80,
            y label style={at={(-0.075,0.5)}},
            x label style={at={(0.5,-0.125)}},
            xtick pos=bottom
        ]
\addplot table[x=interval,y=RL]{\mydatax};
\addplot table[x=interval,y=RND]{\mydatax};
\addplot table[x=interval,y=PMI]{\mydatax};
\addplot table[x=interval,y=PRE]{\mydatax};
\addplot[only marks, mark=starR, black] table[x=interval,y=R]{\mydatax};
\addplot[only marks, mark=starR, black] table[x=interval,y=B]{\mydatax};
\legend{\textit{RL-Profiler}, \textit{RND}+\textit{CNet}, \textit{PMI}+\textit{CNet}, \textit{PT}+\textit{CNet}, \ALL}
\end{axis}
\end{tikzpicture}
\end{subfigure}

\caption{\small{Visual representation of macro \F scores for
    selection-based models with different settings for top-$N$ on validation data.
    The x-axis (not true to scale) shows settings for top-$N$, i.e.,
    $N \in \{5,10,20,30,50\}$ (linearly interpolated), while the y-axis shows
    the corresponding macro \F scores. If $N$ exceeds the number of available 
    posts in profiles, all models converge to the \ALL system since all systems
    select all available posts.}}
\label{graphs_real}
\end{figure}

\subsection{Validation Results on Artificially Enriched Data}
\label{sec:appendix_validation_results_artificial}
\begin{table}[H]
  \centering\small
	\begin{tabular}{lrCCCCCCCCCCCCCCCCCCCCCCCCCCCCCCCC}
\toprule
&& \multicolumn{2}{c}{Open.} & \multicolumn{2}{c}{Consc.} &\multicolumn{2}{c}{Extrav.} &\multicolumn{2}{c}{Agree.} &\multicolumn{2}{c}{Neur.} \\
\cmidrule(rl){3-4} \cmidrule(rl){5-6} \cmidrule(rl){7-8} \cmidrule(rl){9-10}\cmidrule(l){11-12}
System & top-$N$ & m-\F & w-\F & m-\F & w-\F & m-\F & w-\F & m-\F & w-\F & m-\F & w-\F\\
\cmidrule(r){1-1}\cmidrule(rl){2-2}\cmidrule(rl){3-4}\cmidrule(rl){5-6}\cmidrule(rl){7-8}\cmidrule(rl){9-10}\cmidrule(l){11-12}
\RLP & 5 & \BF{100.}\scriptsize{$\pm$0.0}  & \BF{100.}\scriptsize{$\pm$0.0}  & \BF{100.}\scriptsize{$\pm$0.0}  & \BF{100.}\scriptsize{$\pm$0.0}  & \BF{98.1}\scriptsize{$\pm$2.4}  & \BF{98.3}\scriptsize{$\pm$2.2}  & \BF{98.8}\scriptsize{$\pm$1.9}  & \BF{98.9}\scriptsize{$\pm$1.8}  & \BF{97.7}\scriptsize{$\pm$2.5}  & \BF{97.9}\scriptsize{$\pm$2.3} \\
\RND & 5 & 46.8\scriptsize{$\pm$1.6}  & 87.8\scriptsize{$\pm$2.9}  & 37.7\scriptsize{$\pm$5.6}  & 45.0\scriptsize{$\pm$8.0}  & \BF{50.8}\scriptsize{$\pm$10.6}  & \BF{76.5}\scriptsize{$\pm$4.5}  & 59.2\scriptsize{$\pm$7.1}  & 58.4\scriptsize{$\pm$7.4}  & 44.3\scriptsize{$\pm$13.3}  & 42.0\scriptsize{$\pm$15.2} \\
\PMI & 5 & 78.1\scriptsize{$\pm$21.0}  & 93.9\scriptsize{$\pm$5.3}  & \BF{48.9}\scriptsize{$\pm$1.4}  & \BF{60.6}\scriptsize{$\pm$1.8}  & 56.0\scriptsize{$\pm$13.6}  & 80.0\scriptsize{$\pm$5.5}  & 49.3\scriptsize{$\pm$3.6}  & 48.4\scriptsize{$\pm$3.6}  & 40.4\scriptsize{$\pm$5.1}  & 37.7\scriptsize{$\pm$5.3} \\
\PT  & 5 & 70.2\scriptsize{$\pm$24.6}  & 93.7\scriptsize{$\pm$5.0}  & 32.1\scriptsize{$\pm$6.8}  & 40.6\scriptsize{$\pm$7.5}  & 60.0\scriptsize{$\pm$12.3}  & 79.3\scriptsize{$\pm$5.2}  & 67.5\scriptsize{$\pm$4.1}  & 67.8\scriptsize{$\pm$4.1}  & 53.7\scriptsize{$\pm$5.6}  & 53.2\scriptsize{$\pm$6.1} \\
\cmidrule(r){1-1}\cmidrule(rl){2-2}\cmidrule(rl){3-4}\cmidrule(rl){5-6}\cmidrule(rl){7-8}\cmidrule(rl){9-10}\cmidrule(l){11-12}
\RLP & 10 & 100.\scriptsize{$\pm$0.0}  & 100.\scriptsize{$\pm$0.0}  & 78.2\scriptsize{$\pm$6.8}  & 88.4\scriptsize{$\pm$4.5}  & 82.1\scriptsize{$\pm$9.1}  & 89.7\scriptsize{$\pm$5.6}  & 90.7\scriptsize{$\pm$4.5}  & 90.8\scriptsize{$\pm$4.5}  & 93.2\scriptsize{$\pm$4.1}  & 93.8\scriptsize{$\pm$3.7} \\
\RND & 10 & 51.1\scriptsize{$\pm$9.1}  & 87.9\scriptsize{$\pm$2.1}  & 30.1\scriptsize{$\pm$4.9}  & 33.7\scriptsize{$\pm$7.4}  & 49.4\scriptsize{$\pm$11.7}  & 74.1\scriptsize{$\pm$6.5}  & 58.4\scriptsize{$\pm$8.3}  & 57.3\scriptsize{$\pm$8.9}  & 50.0\scriptsize{$\pm$7.8}  & 49.6\scriptsize{$\pm$8.9} \\
\PMI & 10 & \BF{87.8}\scriptsize{$\pm$13.4}  & \BF{96.0}\scriptsize{$\pm$4.5}  & 46.2\scriptsize{$\pm$4.9}  & 56.9\scriptsize{$\pm$6.6}  & 43.6\scriptsize{$\pm$0.6}  & 72.6\scriptsize{$\pm$1.0}  & 56.8\scriptsize{$\pm$6.1}  & 55.5\scriptsize{$\pm$6.7}  & 42.4\scriptsize{$\pm$3.6}  & 39.2\scriptsize{$\pm$4.0} \\
\PT  & 10 & 77.3\scriptsize{$\pm$23.0}  & 94.7\scriptsize{$\pm$5.1}  & 29.1\scriptsize{$\pm$3.6}  & 32.2\scriptsize{$\pm$5.6}  & \BF{81.0}\scriptsize{$\pm$4.4}  & \BF{89.4}\scriptsize{$\pm$2.8}  & \BF{77.6}\scriptsize{$\pm$7.4}  & \BF{77.8}\scriptsize{$\pm$7.4}  & 60.7\scriptsize{$\pm$4.8}  & 60.2\scriptsize{$\pm$5.1} \\
\cmidrule(r){1-1}\cmidrule(rl){2-2}\cmidrule(rl){3-4}\cmidrule(rl){5-6}\cmidrule(rl){7-8}\cmidrule(rl){9-10}\cmidrule(l){11-12}
\RLP & 20 & 94.8\scriptsize{$\pm$16.3}  & 99.1\scriptsize{$\pm$2.9}  & 63.9\scriptsize{$\pm$5.3}  & 77.2\scriptsize{$\pm$5.1}  & 81.5\scriptsize{$\pm$6.1}  & 89.6\scriptsize{$\pm$3.7}  & 82.3\scriptsize{$\pm$2.0}  & 82.2\scriptsize{$\pm$2.1}  & 78.2\scriptsize{$\pm$5.3}  & 81.0\scriptsize{$\pm$4.6} \\
\RND & 20 & 48.0\scriptsize{$\pm$0.7}  & 90.1\scriptsize{$\pm$1.4}  & 37.2\scriptsize{$\pm$4.7}  & 44.4\scriptsize{$\pm$6.9}  & 50.3\scriptsize{$\pm$11.0}  & 75.6\scriptsize{$\pm$5.1}  & 65.0\scriptsize{$\pm$7.1}  & 64.3\scriptsize{$\pm$7.4}  & 57.9\scriptsize{$\pm$6.5}  & 60.7\scriptsize{$\pm$6.5} \\
\PMI & 20 & 48.2\scriptsize{$\pm$0.5}  & 90.4\scriptsize{$\pm$1.0}  & 40.2\scriptsize{$\pm$4.4}  & 48.6\scriptsize{$\pm$6.3}  & 66.5\scriptsize{$\pm$2.8}  & 82.8\scriptsize{$\pm$1.9}  & 66.0\scriptsize{$\pm$4.1}  & 65.1\scriptsize{$\pm$4.3}  & 53.8\scriptsize{$\pm$4.2}  & 54.3\scriptsize{$\pm$4.0} \\
\PT  & 20 & \BF{81.4}\scriptsize{$\pm$16.9}  & \BF{94.3}\scriptsize{$\pm$5.1}  & 37.2\scriptsize{$\pm$5.5}  & 44.3\scriptsize{$\pm$8.3}  & 75.7\scriptsize{$\pm$12.4}  & 86.9\scriptsize{$\pm$6.7}  & 74.8\scriptsize{$\pm$3.8}  & 74.5\scriptsize{$\pm$4.0}  & \BF{69.9}\scriptsize{$\pm$7.8}  & \BF{72.8}\scriptsize{$\pm$6.9} \\
\cmidrule(r){1-1}\cmidrule(rl){2-2}\cmidrule(rl){3-4}\cmidrule(rl){5-6}\cmidrule(rl){7-8}\cmidrule(rl){9-10}\cmidrule(l){11-12}
\RLP & 30 & 84.3\scriptsize{$\pm$25.2}  & 96.9\scriptsize{$\pm$5.1}  & 54.3\scriptsize{$\pm$6.1}  & 67.0\scriptsize{$\pm$7.6}  & 66.7\scriptsize{$\pm$16.8}  & 84.1\scriptsize{$\pm$7.0}  & 76.8\scriptsize{$\pm$3.2}  & 76.6\scriptsize{$\pm$3.3}  & 77.0\scriptsize{$\pm$6.0}  & 80.0\scriptsize{$\pm$5.1} \\
\RND & 30 & 47.9\scriptsize{$\pm$0.8}  & 89.8\scriptsize{$\pm$1.6}  & 38.6\scriptsize{$\pm$7.3}  & 46.1\scriptsize{$\pm$10.5}  & 50.5\scriptsize{$\pm$10.9}  & 75.9\scriptsize{$\pm$4.8}  & 63.8\scriptsize{$\pm$3.7}  & 63.2\scriptsize{$\pm$3.8}  & 56.2\scriptsize{$\pm$7.0}  & 61.0\scriptsize{$\pm$6.7} \\
\PMI & 30 & 48.4\scriptsize{$\pm$0.0}  & 90.7\scriptsize{$\pm$0.0}  & 43.9\scriptsize{$\pm$3.8}  & 53.9\scriptsize{$\pm$5.1}  & \BF{80.1}\scriptsize{$\pm$8.7}  & \BF{90.2}\scriptsize{$\pm$4.0}  & \BF{68.5}\scriptsize{$\pm$4.9}  & \BF{67.8}\scriptsize{$\pm$5.2}  & \BF{57.3}\scriptsize{$\pm$4.3}  & \BF{59.6}\scriptsize{$\pm$4.1} \\
\PT  & 30 & 50.7\scriptsize{$\pm$10.9}  & 89.2\scriptsize{$\pm$2.5}  & 34.8\scriptsize{$\pm$4.4}  & 40.8\scriptsize{$\pm$6.6}  & 68.7\scriptsize{$\pm$5.1}  & 84.0\scriptsize{$\pm$2.7}  & 74.7\scriptsize{$\pm$4.9}  & 74.3\scriptsize{$\pm$5.2}  & 60.4\scriptsize{$\pm$7.7}  & 65.5\scriptsize{$\pm$6.6} \\
\cmidrule(r){1-1}\cmidrule(rl){2-2}\cmidrule(rl){3-4}\cmidrule(rl){5-6}\cmidrule(rl){7-8}\cmidrule(rl){9-10}\cmidrule(l){11-12}
\RLP & 50 & 63.9\scriptsize{$\pm$24.9}  & 93.5\scriptsize{$\pm$4.5}  & 53.0\scriptsize{$\pm$4.2}  & 65.5\scriptsize{$\pm$5.2}  & 50.7\scriptsize{$\pm$11.1}  & 77.9\scriptsize{$\pm$4.5}  & 74.9\scriptsize{$\pm$3.3}  & 74.6\scriptsize{$\pm$3.4}  & 61.5\scriptsize{$\pm$3.5}  & 67.9\scriptsize{$\pm$2.9} \\
\RND & 50 & \BF{53.5}\scriptsize{$\pm$16.3}  & \BF{91.6}\scriptsize{$\pm$2.9}  & \BF{48.0}\scriptsize{$\pm$4.0}  & \BF{59.3}\scriptsize{$\pm$5.3}  & 48.1\scriptsize{$\pm$8.3}  & 76.8\scriptsize{$\pm$3.4}  & \BF{71.7}\scriptsize{$\pm$4.9}  & \BF{71.5}\scriptsize{$\pm$5.0}  & \BF{58.0}\scriptsize{$\pm$7.1}  & \BF{65.1}\scriptsize{$\pm$5.8} \\
\PMI & 50 & 48.4\scriptsize{$\pm$0.0}  & 90.7\scriptsize{$\pm$0.0}  & 47.1\scriptsize{$\pm$4.3}  & 58.1\scriptsize{$\pm$5.7}  & 45.5\scriptsize{$\pm$0.0}  & 75.8\scriptsize{$\pm$0.0}  & 67.1\scriptsize{$\pm$4.9}  & 66.8\scriptsize{$\pm$5.0}  & 53.4\scriptsize{$\pm$6.6}  & 60.7\scriptsize{$\pm$4.8} \\
\PT  & 50 & 58.7\scriptsize{$\pm$21.8}  & 92.6\scriptsize{$\pm$3.9}  & \BF{42.5}\scriptsize{$\pm$4.9}  & \BF{51.8}\scriptsize{$\pm$6.9}  & 56.0\scriptsize{$\pm$13.6}  & 80.0\scriptsize{$\pm$5.5}  & 71.7\scriptsize{$\pm$3.3}  & 71.3\scriptsize{$\pm$3.4}  & 50.8\scriptsize{$\pm$6.4}  & 59.5\scriptsize{$\pm$5.2} \\
\cmidrule(r){1-1}\cmidrule(rl){2-2}\cmidrule(rl){3-4}\cmidrule(rl){5-6}\cmidrule(rl){7-8}\cmidrule(rl){9-10}\cmidrule(l){11-12}
\ALL & all & 89.7\scriptsize{$\pm$21.8}  & 98.2\scriptsize{$\pm$3.9}  & 45.8\scriptsize{$\pm$3.6}  & 63.6\scriptsize{$\pm$3.6}  & 79.1\scriptsize{$\pm$14.2}  & 89.9\scriptsize{$\pm$6.1}  & 77.2\scriptsize{$\pm$4.0}  & 77.7\scriptsize{$\pm$3.9}  & 60.7\scriptsize{$\pm$10.9}  & 67.6\scriptsize{$\pm$8.4} \\
\toprule
	\end{tabular}
  \caption{\small{Macro \F (m-\F) and weighted average \F (w-\F) scores for
    models with different settings for the top-$N$ hyperparameter
    on artificially enriched validation data (averages of 10 runs with standard derivation).
    The best performing setting for top-$N$ (w.r.t. the highest m-\F)
    for each model and personality trait (highlighted in \BF{bold}) is selected
    for evaluation on testing data.}}
  \label{tab:enriched_valid}
\end{table}
\begin{figure}[H]
\centering
\begin{subfigure}{0.5\textwidth}
\pgfplotstableread[row sep=\\,col sep=&]{
  interval & RL    & RND  & PMI   & PRE &R&B\\
5   & 100.0 & 46.8 & 78.1 & 70.2&&\\
10   & 100.0 & 51.1 & 87.8 & 77.3&&\\
20   & 94.8 & 48.0 & 48.2 & 81.4&&\\
30   & 84.3 & 47.9 & 48.4 & 50.7&&\\
50   & 63.9 & 53.5 & 48.4 & 58.7&&\\
All  & 89.7 & 89.7 & 89.7 & 89.7 && 89.7\\
}\mydatax

  \caption*{Openness}
  \vspace{-0.2cm}
\footnotesize
\begin{tikzpicture}
    \begin{axis}[
            width=\widthGraphs,
            height=\heightGraphs,
            legend style={at={(1.32,0)},anchor=south east},
            symbolic x coords={5,10,.,20,.,30,.,.,50,.,.,All},
            ymajorgrids=true,
            xtick=data,
            every axis plot/.append style={semithick},
            nodes near coords align={vertical},
            ymin=25,ymax=100,
            y label style={at={(-0.075,0.5)}},
            x label style={at={(0.5,-0.125)}},
            xtick pos=bottom
        ]
\addplot table[x=interval,y=RL]{\mydatax};
\addplot table[x=interval,y=RND]{\mydatax};
\addplot table[x=interval,y=PMI]{\mydatax};
\addplot table[x=interval,y=PRE]{\mydatax};
\addplot[only marks, mark=starR, black] table[x=interval,y=R]{\mydatax};
\addplot[only marks, mark=starR, black] table[x=interval,y=B]{\mydatax};
\end{axis}
\end{tikzpicture}
\end{subfigure}
\hspace{-0.35cm}
\begin{subfigure}[b]{0.5\textwidth}
\pgfplotstableread[row sep=\\,col sep=&]{
  interval & RL    & RND  & PMI   & PRE &R&B\\
5   & 100.0 & 37.7 & 48.9 & 32.1&&\\
10   & 78.2 & 30.1 & 46.2 & 29.1&&\\
20   & 63.9 & 37.2 & 40.2 & 37.2&&\\
30   & 54.3 & 38.6 & 43.9 & 34.8&&\\
50   & 53.0 & 48.0 & 47.1 & 42.5&&\\
All  & 45.8 & 45.8 & 45.8 & 45.8 && 45.8\\
    }\mydata

  \caption*{Conscientiousness}
\vspace{-0.2cm}
\footnotesize
\begin{tikzpicture}
    \begin{axis}[
            width=\widthGraphs,
            height=\heightGraphs,
            legend style={at={(1.32,0)},anchor=south east},
            symbolic x coords={5,10,.,20,.,30,.,.,50,.,.,All},
            ymajorgrids=true,
            xtick=data,
            every axis plot/.append style={semithick},
            nodes near coords align={vertical},
            ymin=25,ymax=100,
            y label style={at={(-0.075,0.5)}},
            x label style={at={(0.5,-0.15)}},
            xtick pos=bottom
        ]
\addplot table[x=interval,y=RL]{\mydata};
\addplot table[x=interval,y=RND]{\mydata};
\addplot table[x=interval,y=PMI]{\mydata};
\addplot table[x=interval,y=PRE]{\mydata};
\addplot[only marks, mark=starR, black] table[x=interval,y=R]{\mydata};
\addplot[only marks, mark=starR, black] table[x=interval,y=B]{\mydata};
\end{axis}
\end{tikzpicture}
\end{subfigure}

\begin{subfigure}{0.5\textwidth}
\pgfplotstableread[row sep=\\,col sep=&]{
  interval & RL    & RND  & PMI   & PRE &R&B\\
5   & 98.1 & 50.8 & 56.0 & 60.0&&\\
10   & 82.1 & 49.4 & 43.6 & 81.0&&\\
20   & 81.5 & 50.3 & 66.5 & 75.7&&\\
30   & 66.7 & 50.5 & 80.1 & 68.7&&\\
50   & 50.7 & 48.1 & 45.5 & 56.0&&\\
All  & 79.1 & 79.1 & 79.1 & 79.1 && 79.1\\
}\mydatax

  \caption*{Extraversion}
  \vspace{-0.2cm}
\footnotesize
\begin{tikzpicture}
    \begin{axis}[
            width=\widthGraphs,
            height=\heightGraphs,
            legend style={at={(1.32,0)},anchor=south east},
            symbolic x coords={5,10,.,20,.,30,.,.,50,.,.,All},
            ymajorgrids=true,
            xtick=data,
            every axis plot/.append style={semithick},
            nodes near coords align={vertical},
            ymin=25,ymax=100,
            y label style={at={(-0.075,0.5)}},
            x label style={at={(0.5,-0.125)}},
            xtick pos=bottom
        ]
\addplot table[x=interval,y=RL]{\mydatax};
\addplot table[x=interval,y=RND]{\mydatax};
\addplot table[x=interval,y=PMI]{\mydatax};
\addplot table[x=interval,y=PRE]{\mydatax};
\addplot[only marks, mark=starR, black] table[x=interval,y=R]{\mydatax};
\addplot[only marks, mark=starR, black] table[x=interval,y=B]{\mydatax};
\end{axis}
\end{tikzpicture}
\end{subfigure}
\hspace{-0.35cm}
\begin{subfigure}[b]{0.5\textwidth}
\pgfplotstableread[row sep=\\,col sep=&]{
  interval & RL    & RND  & PMI   & PRE &R&B\\
5   & 98.8 & 59.2 & 49.3 & 67.5&&\\
10   & 90.7 & 58.4 & 56.8 & 77.6&&\\
20   & 82.3 & 65.0 & 66.0 & 74.8&&\\
30   & 76.8 & 63.8 & 68.5 & 74.7&&\\
50   & 74.9 & 71.7 & 67.1 & 71.7&&\\
All  & 77.2 & 77.2 & 77.2 & 77.2 && 77.2\\
    }\mydata

  \caption*{Agreeableness}
\vspace{-0.2cm}
\footnotesize
\begin{tikzpicture}
    \begin{axis}[
            width=\widthGraphs,
            height=\heightGraphs,
            legend style={at={(1.32,0)},anchor=south east},
            symbolic x coords={5,10,.,20,.,30,.,.,50,.,.,All},
            ymajorgrids=true,
            xtick=data,
            every axis plot/.append style={semithick},
            nodes near coords align={vertical},
            ymin=25,ymax=100,
            y label style={at={(-0.075,0.5)}},
            x label style={at={(0.5,-0.15)}},
            xtick pos=bottom
        ]
\addplot table[x=interval,y=RL]{\mydata};
\addplot table[x=interval,y=RND]{\mydata};
\addplot table[x=interval,y=PMI]{\mydata};
\addplot table[x=interval,y=PRE]{\mydata};
\addplot[only marks, mark=starR, black] table[x=interval,y=R]{\mydata};
\addplot[only marks, mark=starR, black] table[x=interval,y=B]{\mydata};
\end{axis}
\end{tikzpicture}
\end{subfigure}

\begin{subfigure}{0.5\textwidth}
\pgfplotstableread[row sep=\\,col sep=&]{
  interval & RL    & RND  & PMI   & PRE &R&B\\
5   & 97.7 & 44.3 & 40.4 & 53.7&&\\
10   & 93.2 & 50.0 & 42.4 & 60.7&&\\
20   & 78.2 & 57.9 & 53.8 & 69.9&&\\
30   & 77.0 & 56.2 & 57.3 & 60.4&&\\
50   & 61.5 & 58.0 & 53.4 & 50.8&&\\
All  & 60.7 & 60.7 & 60.7 & 60.7 && 60.7\\
}\mydatax

  \caption*{Neuroticism}
  \vspace{-0.2cm}
\footnotesize
\begin{tikzpicture}
    \begin{axis}[
            width=\widthGraphs,
            height=\heightGraphs,
            legend style={at={(1.61,0)},anchor=south east},
            symbolic x coords={5,10,.,20,.,30,.,.,50,.,.,All},
            ymajorgrids=true,
            xtick=data,
            every axis plot/.append style={semithick},
            nodes near coords align={vertical},
            ymin=25,ymax=100,
            y label style={at={(-0.075,0.5)}},
            x label style={at={(0.5,-0.125)}},
            xtick pos=bottom
        ]
\addplot table[x=interval,y=RL]{\mydatax};
\addplot table[x=interval,y=RND]{\mydatax};
\addplot table[x=interval,y=PMI]{\mydatax};
\addplot table[x=interval,y=PRE]{\mydatax};
\addplot[only marks, mark=starR, black] table[x=interval,y=R]{\mydatax};
\addplot[only marks, mark=starR, black] table[x=interval,y=B]{\mydatax};
\legend{\RLP, \textit{RND}+\textit{CNet}, \textit{PMI}+\textit{CNet}, \textit{PT}+\textit{CNet}, \ALL}
\end{axis}
\end{tikzpicture}
\end{subfigure}
\caption{\small{Visual representation of macro \F scores for
    selection-based models with different settings for top-$N$ on artificially enriched validation data.
    The x-axis (not true to scale) shows settings for top-$N$, i.e.,
    $N \in \{5,10,20,30,50\}$ (linearly interpolated), while the y-axis shows
    the corresponding macro \F scores. If $N$ exceeds the number of available 
    posts in profiles, all models converge to the \ALL system since all systems
    select all available posts.}}
\label{graphs_enriched}
\end{figure}

\newpage
\subsection{Artificial Post Generation and Dataset Enrichment}
\label{sec:appendix_data_generation}
\begin{figure*}[t]
  \footnotesize
	\begin{Verbatim}[frame=single]
Recall the personality trait extraversion.
A person with a high level of extraversion may see themselves as someone who is talkative, or {...}
Generate ten tweets that are likely written by a person with a high level of extraversion.
+Do not use emojis or hashtags. Try to include the topic {topic}.
	\end{Verbatim}
	\begin{Verbatim}[frame=single]
Recall the personality trait extraversion.
A person with a low level of extraversion may see themselves as someone who is reserved, or {...}
Generate ten tweets that are likely written by a person with a low level of extraversion.
+Do not use emojis or hashtags. Try to include the topic {topic}.
  \end{Verbatim}
	\vspace{-0.35cm}
\caption{Prompt templates for generating artificial posts indicating a \textit{high} and \textit{low} level of \extraversion.}
  \label{verb:prompt_template_gen}
\end{figure*}

\noindent
To generate artificial posts indicating either a \textit{low} or
\textit{high} level of a certain personality trait we use
Llama~2~13B-Chat, and repeatedly prompt the model to generate 10 posts.
We present the prompt templates we use for generating artificial posts for the
\extraversion trait in Figure~\ref{verb:prompt_template_gen}. Here, the task
of generating posts is verbalized by the phrase ``Generate ten tweets that are
likely written by a person with a high level of extraversion''. Similarly to the
prompts used in CNet for prediction levels of a trait, we include BFI-44
items to enrich context. The prompts for the other Big Five traits follow a similar
structure.

To further encourage diversity in the generating posts, since different
profiles should be enriched with different posts (because it would be trivial
for a model to find these posts if they are always the same), we task the LLM
to include a topic in the generated posts. For this, we compile a list of 12
topics we derive from the work by \citet{antypas-etal-2022-twitter} covering
many discussion points in social media:
\begin{minipage}[t]{0.49\textwidth}
\begin{itemize}
  \setlength\itemsep{-0.5em}
  \item News
  \item Social Concern
  \item Sports
  \item Music
  \item Celebrity {\&} Pop Culture
  \item Film, TV {\&} Video
\end{itemize}
\end{minipage}
\begin{minipage}[t]{0.49\textwidth}
\begin{itemize}
  \setlength\itemsep{-0.5em}
  \item Diaries {\&} Daily Life
  \item Arts {\&} Culture
  \item Science {\&} Technology
  \item Fitness {\&} Health
  \item Family
  \item Relationships
\end{itemize}
\end{minipage}
\vspace{0.3cm}

\noindent
We present examples of artificially generated posts for different topics
and personality traits in Table~\ref{tab:prompt_template_gen_examples}.

We use the artificially generated posts in our post-hoc analysis. Here, we draw
subsets from the datasets we derived from the PAN-AP-2015 corpus. For each trait,
split and class we randomly select 15 profiles. Note that in some partitions, there
are less than 15 profiles and this process therefore changes class distribution in each
sub-corpus. Table~\ref{table:PAN2015_statistics_small} shows statistics of the
datasets we obtain by this. Afterward, we enrich each profile in these
dataset splits with five artificially generated posts we randomly choose from
the pool of generated posts (ensuring we use each artificial post only once) based
on their ground-truth annotation, e.g., for profiles annotated with a
\textit{low} level of \extraversion, we add generated posts that aim to
indicate a \textit{low} level of \extraversion.

\begin{table*}[t]
	\centering\small
	\begin{tabular}{lrrrrrrrrrrrrrrrrr}
		\toprule
		       & \multicolumn{2}{c}{Training} & \multicolumn{2}{c}{Validation} & \multicolumn{2}{c}{Testing} \\
		\cmidrule(rl){2-3}\cmidrule(rl){4-5}\cmidrule(rl){6-7}
    Class  & \multicolumn{1}{c}{High}     & \multicolumn{1}{c}{Low}    & \multicolumn{1}{c}{High}     & \multicolumn{1}{c}{Low}      & \multicolumn{1}{c}{High}    & \multicolumn{1}{c}{Low} \\
		\cmidrule(r){1-1}\cmidrule(rl){2-3}\cmidrule(rl){4-5}\cmidrule(rl){6-7}
		Openness  & 15 & 1 & 15 & 1 & 15 & 1   \\
		Conscientiousness & 15 & 3 & 15 & 2 & 15 & 10  \\
		Extraversion & 15 & 12& 15 & 3 & 15 & 6   \\
		Agreeableness & 15 & 15 & 15 & 4 & 15 & 11 \\
		Neuroticism & 15 & 15 & 15 & 8 & 15 & 15 \\
		\bottomrule
	\end{tabular}
  \caption{Corpora statistics of the splits derived from the
    PAN-AP-2015 \citep{rangel2015overview} corpus for post-hoc
    experiments on partially artificially data (in numbers of profiles).}
	\label{table:PAN2015_statistics_small}
\end{table*}

\begin{table}[p!]
\centering\small
\begin{tabular}{L{0.1cm}L{0.1cm}ll}
\toprule
&& Topic & Generated Post \\
\cmidrule(rl){3-3}\cmidrule(l){4-4}
\multirow{10}{*}{\rotatebox[origin=l]{90}{Openness}} & \multirow{5}{*}{\rotatebox[origin=l]{90}{High}}
    & News & I'm not just a thinker, I'm a dreamer. And I believe that dreams can change the world.\\
    && Social Concern & Why are we still tolerating systemic injustices? It's time for change.\\
    && Music & I love how music can transport me to another world, even if just for a few minutes\\
    && Science {\&} Techn. & Building a DIY robot using an Arduino board and exploring its capabilities.\\
    && Family & Family is a source of comfort and strength, especially during challenging times.\\
    \cmidrule(rl){3-3}\cmidrule(l){4-4}
    & \multirow{5}{*}{\rotatebox[origin=l]{90}{Low}}
    & Social Concern & I don't care about the homeless crisis in our city, it's not my problem.\\
    && Sports & Just finished a 9-to-5 shift at the office and now I'm ready to watch some football!\\
    && Music & Classical music is the only real music. Everything else is just noise.\\
    && Arts {\&} Culture & I don't read books. I don't see the point of wasting time on made-up stories.\\
    && Science {\&}  Techn. & I don't see the point of space exploration. We have enough problems on Earth\\
\cmidrule{1-4}

\multirow{10}{*}{\rotatebox[origin=l]{90}{Conscientiousness}} & \multirow{5}{*}{\rotatebox[origin=l]{90}{High}}
    & Sports & I strive for efficiency in everything I do, whether it's on the field or in the weight room.\\
    && Music & I've been practicing my guitar for hours every day to perfect my technique.\\
    && Film, TV {\&}  Video& I'm so impressed by the cinematography in the latest blockbuster. It's like a work of art.\\
    && Diaries {\&}  Daily Life& I find solace in my daily routine, it brings me a sense of stability and predictability.\\
    && Fitness {\&}  Health & I track my progress and adjust my plan as needed to ensure I'm reaching my fitness goals.\\
    \cmidrule(rl){3-3}\cmidrule(l){4-4}

    & \multirow{5}{*}{\rotatebox[origin=l]{90}{Low}}
    & News & Can't find my homework... or my textbook... or my notes. Anyone have a photocopy?\\
    && Sports & I think I might have accidentally signed up for a relay instead of a solo race\\
    && Film, TV {\&}  Video & I'm so addicted to my favorite TV show that I can't stop thinking about it.  I need help! \\
    && Diaries {\&}  Daily Life & I just spent \$100 on a new outfit instead of paying my rent. Oopsie.\\
    && Relationships & I know I said I would call my partner back yesterday, but uh... I forgot?\\
\cmidrule{1-4}

\multirow{10}{*}{\rotatebox[origin=l]{90}{Extraversion}} & \multirow{5}{*}{\rotatebox[origin=l]{90}{High}}
    & News & I'm so excited to share the latest scoop with all my followers!\\
    && Music & Just discovered a new artist and I can't stop listening to their music!\\
    && Diaries {\&}  Daily Life & I just tried the craziest new food trend and it was so good! I can't wait to try more\\
    && Fitness {\&}  Health & Feeling so strong and confident after a killer leg day at the gym.\\
    && Relationships & I'm not scared of rejection. I'll put myself out there and see what happens!\\

    \cmidrule(rl){3-3}\cmidrule(l){4-4}
    & \multirow{5}{*}{\rotatebox[origin=l]{90}{Low}}
    & Sports & I prefer to focus on my own improvement rather than comparing myself to others.\\
    && Music & My favorite way to relax is to listen to calming music and meditate.\\
    && Science {\&}  Techn. & My mind is always racing with ideas, but I struggle to express them out loud.\\
    && Fitness {\&}  Health & I'm not a fan of loud, crowded gyms, I prefer to work out at home in my own space.\\
    && Family & I love my family, but sometimes I just need a little alone time to recharge.\\
\cmidrule{1-4}

\multirow{10}{*}{\rotatebox[origin=l]{90}{Agreeableness}} & \multirow{5}{*}{\rotatebox[origin=l]{90}{High}}
    & Social Concern & I'm a team player, and I think collaboration is the key to success. \\
    && Sports & I can't believe we won!  It's all thanks to our teamwork and determination.\\
    && Diaries {\&}  Daily Life& I think it's important to be open-minded and accepting of others.\\
    && Fitness {\&}  Health& I'm so grateful for my fitness community - they inspire me to be my best self every day.\\
    && Family & I love being a part of our family's traditions and making new memories together.\\

    \cmidrule(rl){3-3}\cmidrule(l){4-4}
    & \multirow{5}{*}{\rotatebox[origin=l]{90}{Low}}
    & News & I can't believe the media is still covering that story, it's such a non-issue.\\
    && Social Concern & I don't have time for weak people, they need to toughen up.\\
    && Sports & Why should I have to follow the rules? The other team is always cheating anyway.\\
    && Science {\&}  Techn. & Technology is ruining our society. We need to go back to simpler times.\\
    && Family & My family is always trying to tell me what to do. Newsflash: I don't need their advice.\\
\cmidrule{1-4}

\multirow{10}{*}{\rotatebox[origin=l]{90}{Neuroticism}} & \multirow{5}{*}{\rotatebox[origin=l]{90}{High}}
    & News & I can't believe what I just heard on the news. It's like, what is even happening?!\\
    && Sports & I'm so tense before every game. I can't relax, no matter how hard I try.\\
    && Diaries {\&}  Daily Life & I've been doing yoga for months and still can't touch my toes. \\
    && Arts {\&} Culture & Why can't I just enjoy a simple painting without overanalyzing every brushstroke? \\
    && Family & My family is always causing drama. I just want peace and quiet!\\

    \cmidrule(rl){3-3}\cmidrule(l){4-4}
    & \multirow{5}{*}{\rotatebox[origin=l]{90}{Low}}
    & Social Concern & I'm not perfect, but I strive to be a good listener and a supportive friend.\\
    && Celebr. {\&}  Pop Cult. & I don't stress about fashion or beauty trends. Comfort and  simplicity are key for me!\\
    && Diaries {\&}  Daily Life& I'm proud of my ability to remain emotionally stable, even in difficult situations.\\
    && Arts {\&} Culture & The beauty of nature is a never-ending source of inspiration for my art. \\
    && Family & Family vacations are the best kind of stress-free fun. \\
\bottomrule
\end{tabular}
\caption{Examples of posts generated using Llama 2 13B-Chat that aim to indicate
either a \textit{low} or \textit{high} level of one of the Big Five traits.}
\label{tab:prompt_template_gen_examples}
\end{table}

\end{document}